\documentclass[times,twocolumn,final]{elsarticle}
\usepackage{medima}
\usepackage{framed,multirow}

\usepackage{amssymb}
\usepackage{latexsym}
\usepackage{url}
\usepackage[dvipsnames]{xcolor}
\usepackage{hyperref}

\usepackage{amsmath,amssymb,amsfonts}
\usepackage{breqn}

\usepackage{algorithm}% http://ctan.org/pkg/algorithms
\usepackage{algpseudocode}% http://ctan.org/pkg/algorithmicx

\usepackage{graphicx}
\usepackage{textcomp}
\usepackage{thmtools, thm-restate}

\usepackage{dirtytalk}
\usepackage{epsfig}
\usepackage{booktabs}
\usepackage{verbatim}
\usepackage{xcolor}
\usepackage{amstext}
\usepackage{upgreek}
\usepackage{multirow}
\usepackage{bbm}
\usepackage{rotating}
\usepackage{pifont}
\usepackage{colortbl}
\usepackage{arydshln}
\usepackage{subfig}
\usepackage{amsfonts}
\usepackage{threeparttable, tablefootnote}
\usepackage{tabularray}
\usepackage{transparent}

\def\rmX{{\mathbf{X}}}
\def\rmY{{\mathbf{Y}}}
\def\rmZ{{\mathbf{Z}}}
\def\rmw{{\mathbf{w}}}
\def\real{{\mathbb{R}}}

\definecolor{mypink1}{rgb}{0.858, 0.188, 0.478}
\definecolor{myorange}{RGB}{255,165,0}
\definecolor{gray}{cmyk}{0.86,0.86,0.86,0.86}
\usepackage{color, colortbl}
\definecolor{Gray}{gray}{0.9}
\definecolor{newcolor}{rgb}{.8,.349,.1}

\newcommand{\our}{\cellcolor{Gray}}
\newcommand{\improvement}[1]{$_{(}$\textcolor{blue}{$_{+#1}$}$_{)}$}
\newcommand{\worsening}[1]{$_{(}$\textcolor{red}{$_{-#1}$}$_{)}$}

\usepackage[switch]{lineno}

\makeatletter
\def\adl@drawiv#1#2#3{%
        \hskip.5\tabcolsep
        \xleaders#3{#2.5\@tempdimb #1{1}#2.5\@tempdimb}%
                #2\z@ plus1fil minus1fil\relax
        \hskip.5\tabcolsep}
\newcommand{\cdashlinelr}[1]{%
  \noalign{
           \global\let\@dashdrawstore\adl@draw
           \global\let\adl@draw\adl@drawiv}
  \cdashline{#1}
  \noalign{\global\let\adl@draw\@dashdrawstore
           }}
\makeatother

\makeatletter
\newcommand\notsotiny{\@setfontsize\notsotiny\@vipt\@viipt}
\makeatother

\journal{Medical Image Analysis}

\begin{document}

\verso{Silva-Rodríguez \textit{et~al.}}

\begin{frontmatter}

\title{Towards Foundation Models and Few-Shot Parameter-Efficient Fine-Tuning for Volumetric Organ Segmentation}

\author[1]{Julio Silva-Rodríguez$^\ast$}
\cortext[cor1]{Corresponding author: julio-jose.silva-rodriguez@etsmtl.ca}
\author[1,3]{Jose Dolz}
\author[1,3]{Ismail {Ben Ayed}}

\address[1]{ÉTS Montréal, Québec, Canada}
\address[3]{Centre de Recherche du Centre Hospitalier de l’Universit\'e de Montr\'eal (CR-CHUM), Québec, Canada}

%\received{1 May 2013}
%\finalform{10 May 2013}
%\accepted{13 May 2013}
%\availableonline{15 May 2013}
%\communicated{S. Sarkar}

%% Abstract
%-----------
\begin{abstract}
The recent popularity of foundation models and the pre-train-and-adapt paradigm, where a large-scale model is transferred to downstream tasks, is gaining attention for volumetric medical image segmentation. However, current transfer learning strategies devoted to full fine-tuning for transfer learning may require significant resources and yield sub-optimal results when the labeled data of the target task is scarce. This makes its applicability in real clinical settings challenging since these institutions are usually constrained on data and computational resources to develop proprietary solutions. To address this challenge, we formalize Few-Shot Efficient Fine-Tuning (FSEFT), a novel and realistic scenario for adapting medical image segmentation foundation models. This setting considers the key role of both data- and parameter-efficiency during adaptation. Building on a foundation model pre-trained on open-access CT organ segmentation sources, we propose leveraging Parameter-Efficient Fine-Tuning and black-box Adapters to address such challenges. Furthermore, novel efficient adaptation methodologies are introduced in this work, which include Spatial black-box Adapters that are more appropriate for dense prediction tasks and constrained transductive inference, leveraging task-specific prior knowledge. Our comprehensive transfer learning experiments confirm the suitability of foundation models in medical image segmentation and unveil the limitations of popular fine-tuning strategies in few-shot scenarios. The project code is available: \url{https://github.com/jusiro/fewshot-finetuning}. \\


\end{abstract}

\end{frontmatter}

%\linenumbers

%% Introduction
%-----------
\section{Introduction}
\label{sec:intro}

The recent advancements in deep learning have yielded remarkable outcomes in visual recognition tasks. Specifically, training on large enough amounts of labeled data provides excellent medical image segmentation performance under the standard supervised learning paradigm. The success of several recent public challenges, including \cite{MSD, AMOS, KiTS}, attests to this. However, these models are often trained on a specific task and limited numbers of samples, which may lack real-world inter-center variability. As a result, the current literature suggests that general medical image segmentation is hampered by the lack of large, curated datasets for training \citep{Chen2022}. This limitation is further exacerbated in volumetric medical image segmentation, where expert knowledge is required for voxel-wise annotations. For example, an experienced clinician would require an average of 10 minutes to segment a unique structure in a CT scan \citep{TotalSeg}. Recent works in the medical image analysis literature have examined the potential of large-scale \textit{foundation models} in organ segmentation by integrating multiple publicly available datasets corresponding to various tasks \citep{UniversalModel,ye2023uniseg,multitalent}. The emerging idea of building foundation models for medical image segmentation follows from the current substantial paradigm shift in computer vision and natural language processing. This is driven by the growing prevalence of large models \citep{Brown-2020, radford2021learning} trained on diverse datasets. Such models have shown outstanding generalization capabilities and can be adapted to a breadth of target tasks using only a few labeled samples in the target categories, also known as \textit{few-shot adaptation}. For instance, vision-language models, such as CLIP (Contrastive Language-Image Pre-training) \citep{radford2021learning}, are transforming computer vision, emerging as a promising solution towards true generalization. Such models yield robust features, providing powerful alternatives to standard dataset-specific models trained on orders of magnitude lower amounts of data. For example, they have already shown very promising few-shot adaptation in the context of image classification tasks \citep{Zhang2022,Gao2021}. Also, the recent foundation SAM (`'Segment Anything'' Model) \citep{Kirillov-2023}, built for segmentation with over 11M images and 1B masks, has attracted broad interest and yielded impressive performances on natural images.

\begin{figure*}[t!]
    \centering

            \begin{tabular}{ccc}
            \includegraphics[width=.27\linewidth]{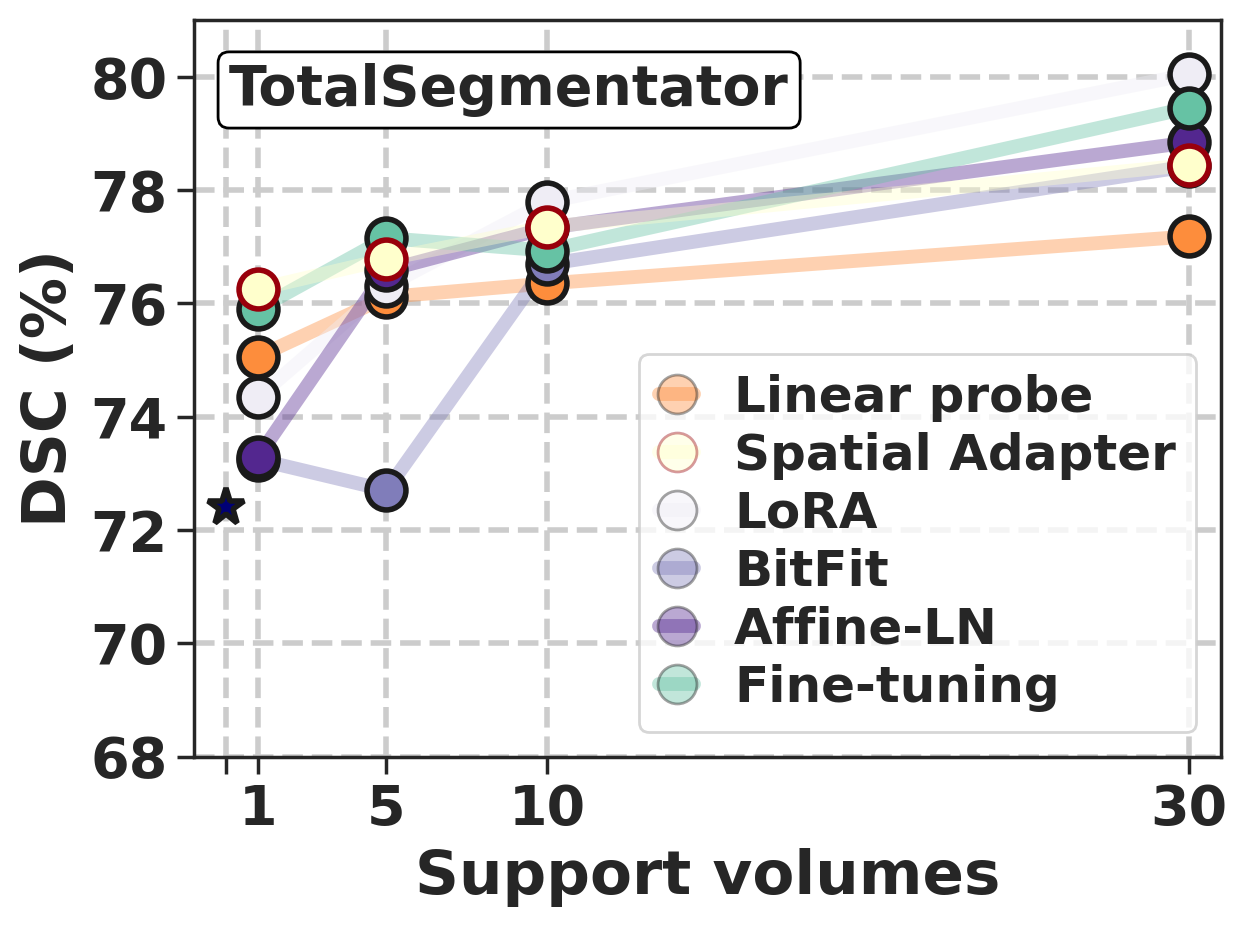} &
            \includegraphics[width=.27\linewidth]{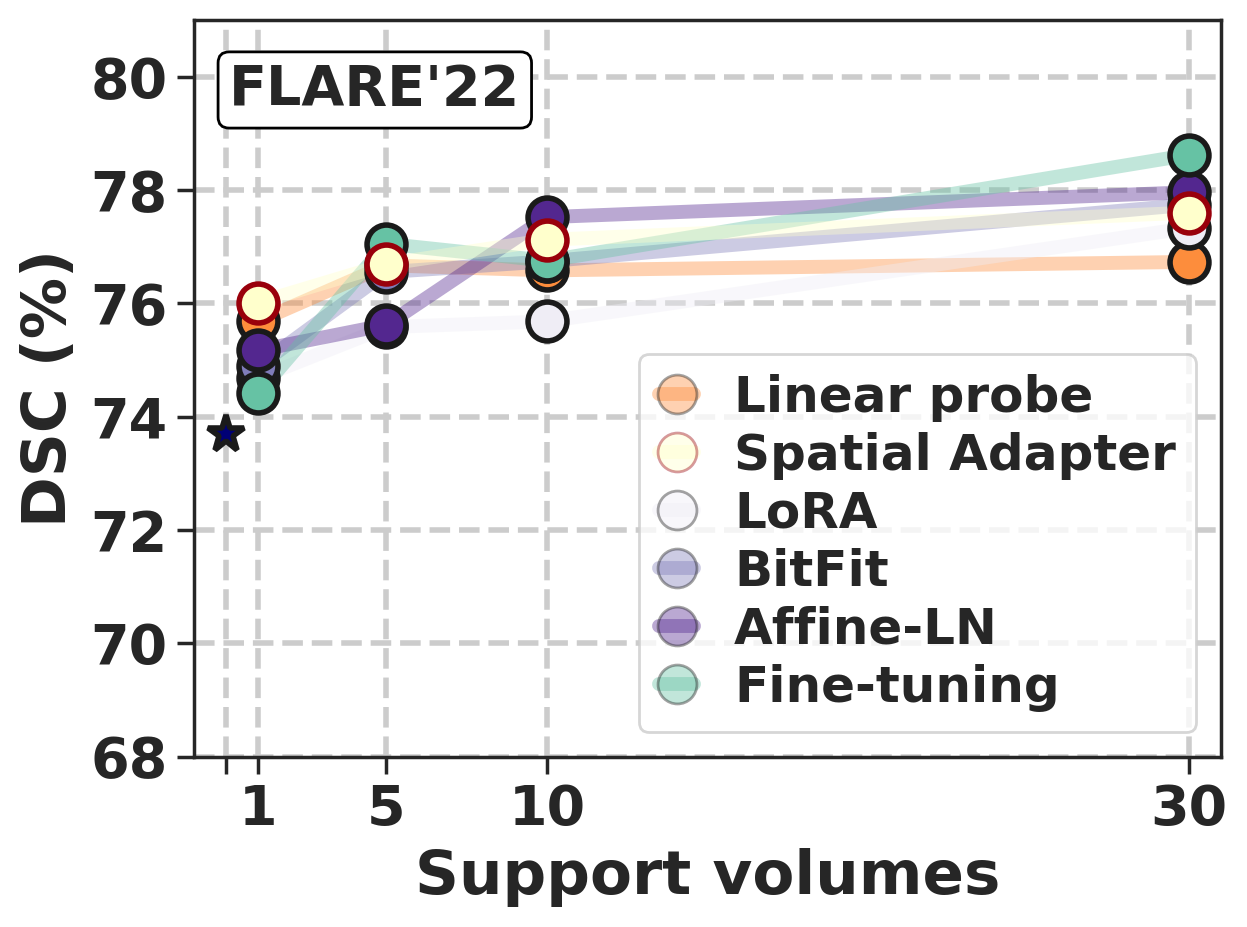} &
            \includegraphics[width=.27\linewidth]{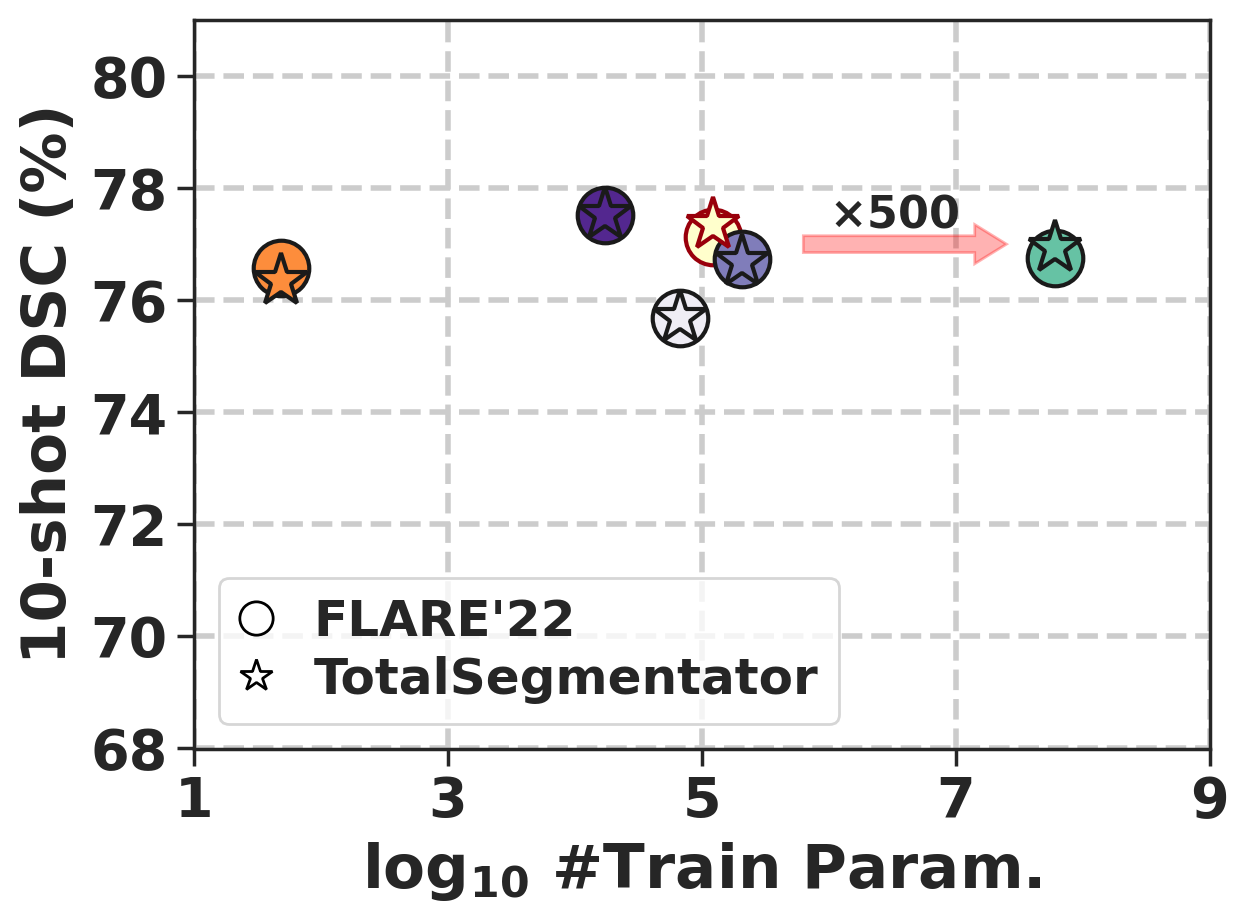} \\
            \multicolumn{2}{c}{\small \textbf{(a) Data-efficient adaptation}} & \small \textbf{(b) Parameter efficiency}
             \end{tabular}
             
        \caption{\textbf{Towards efficiently adapting medical volumetric foundation models.} In this work, we introduce a foundation model for volumetric CT organ segmentation, pre-trained on 2,042 partially annotated scans. We then delve into clinically realistic settings to adapt these models, considering \textbf{(a) \textit{data}} and \textbf{(b) \textit{parameter efficiency}}. These motivations formalize our proposed transfer learning setting, \textbf{Few-Shot Efficient Fine-Tuning (FSEFT)}, which leverages popular Parameter-Efficient Fine-Tuning methods and novel privacy-preserving black-box Adapters to address such real-world challenges. DSC: Dice similarity coefficient.}
        \label{fig:intro}
\end{figure*}

Alongside the establishment of foundation models as a cornerstone in computer vision and natural language processing applications, the medical imaging community is transitioning from narrowly supervised, task-specific models with limited generalization capabilities towards the new pre-train-and-adapt paradigm \citep{Moor-Nature-2023}. Indeed, there is currently emergent literature on foundation classification or segmentation models, which are tailored to specific clinical domains, including radiology \citep{Zhang-23-NatureCom,Wang-NeurIPS-2022}, ophthalmology \citep{Zhou-Nature-2023,Silva-Rodriguez-23-MedIA}, histology \citep{Huang-NatureMed-2023,Zhang-MedAGI-2023}, and endoscopy \citep{Wang-MICCAI-2023}, among others. Many of these recent works pointed to the limited performances of generalist foundation models (like CLIP or SAM) in medical tasks \citep{Mazurowski-2023,Zhang-23-NatureCom,Silva-Rodriguez-23-MedIA}, as they may not capture specialized medical-domain knowledge and its fine-grained features. This calls for developing specialized pre-training and adaptation methodologies, which account for the complexity of medical imaging data and the associated expert domain knowledge. 

Most of these recent medical imaging foundation models have been investigated for image classification tasks. Indeed, despite the strong potential of foundation models, the literature on this subject for medical image segmentation is still scarce, with only a few very recent works, e.g., \cite{UniversalModel,multitalent,ye2023uniseg}. Furthermore, these works assume that a large set of labeled samples is accessible during the adaptation of such networks to the new tasks/domains. In the medical context, however, since each institute has limited time, budget, and particular clinical purposes, the number of annotated samples available in clinical practice is usually limited. Therefore, few-shot adaptation, which assumes access to only a few labeled samples, is very appealing in medical image segmentation. This motivates the development of new paradigms that allow for data-efficient adaptation of foundation models in this field. It is worth mentioning that, in computer vision, there is currently a substantial interest in adapting foundation image-language models in few-shot regimes \cite{Zhou-22, Zhang2022}, primarily focusing on natural image classification. More particularly, black-box Adapters operating over pre-trained features have been recently popularized in vision-language models to tackle this challenge \citep{Gao2021, Zhang2022, CLAP}.

In addition, we observe that such concurrent works developing specialized segmentation foundation models \cite{UniversalModel,multitalent,suprem} usually transfer the pre-trained model by fully fine-tuning its weights to tackle the new task, which has notorious limitations. First, modern deep-learning models typically are largely parameterized networks and require substantial hardware requirements for training, which may be unsustainable in clinical institutions. In addition, storing a different model for each new domain/task might be expensive, especially considering the number of parameters of the largest-scale state-of-the-art Transformers-based 3D segmentation networks such as UNETR (555M) \citep{UNETR} or SwinUNETR (371M) \citep{hatamizadeh2022swin}. Second, according to recent literature on image classification with vision-language models, fine-tuning might distort the pre-trained representation and not generalize properly when adapted \citep{LPFT}. This problem is commonly known in the machine learning literature as catastrophic forgetting \citep{catforg} and acquires paramount importance in the upcoming era of foundation models. To address both issues, the natural language processing community has recently devoted extensive efforts to Parameter-Efficient Fine-Tuning (PEFT), i.e., how to adapt a large pre-trained model to newly collected data in a computationally efficient way. In contrast to full fine-tuning, PEFT involves tuning a small subset of the existing model layers and/or adding a new, limited set of parameters, so-called Adapters. The impact of PEFT in this context goes without saying. Currently, state-of-the-art large language models can be adapted with commodity hardware devices using, for example, the popular Low-Rank Adapters, a.k.a. LoRA \citep{hu2022lora}. However, its impact on medical image analysis is yet to be determined \citep{Dutt-missing-opportunity-23}.

Based on these considerations of the current state of medical image segmentation foundation models, we ought to direct the efforts toward more efficient and realistic transfer learning strategies to further enhance their application in clinical settings. Concretely, we provide the following contributions:

\begin{itemize}

     \item A foundation model for volumetric organ segmentation is released, trained on nine publicly available datasets gathering 2,042 CT scans and 29 annotated structures.

     \item We formalize Few-Shot Efficient Fine-Tuning (FSEFT), a novel and realistic setting for transferring supervised pre-trained 3D models in challenging clinical scenarios. Thus, FSEFT considers the scarcity of adaptation supervision, using only a handful of labeled samples in the target task, and the parameter efficiency. 

     \item Comprehensive transferability experiments on external datasets are reported, exploring the transfer learning capabilities of popular PEFT methods and black-box Adapters in segmenting known and novel organs under severe domain drifts.

    \item Additionally, we introduce different technical contributions to go further in such adaptation, which include (a) black-box Spatial Adapters tailored to dense-prediction tasks; (b) a constrained transductive inference, which leverages task-specific prior knowledge; (c) investigating the importance of classification head initialization for improved transfer in low-data regimes.

    \item We provide empirical insights that point out the potential of PEFT and black-box Adapters compared to the standard fine-tuning while requiring a fraction of tuning parameters. A summary of the FSEFT capabilities is introduced in Figure~\ref{fig:intro}. Beyond our foundation model, the proposed setting is universally applicable to any supervised pre-trained foundation model, as we empirically demonstrate using recently released models from other authors \citep{UniversalModel,suprem}. These results highlight the benefits of our framework in practical clinical settings.

\end{itemize}

%% Related work
%-----------
\section{Related Work}
\label{sec:rw}

\subsection{Pre-training transferable medical volumetric models}

Foundation models require a large neural network architecture and enough data to pre-train a rich, transferable feature representation on the data domain. Two paradigms constantly compete to address this task: self-supervised and supervised pre-training. The first family of methods aims to learn such representations from unlabeled data, thus reducing the need for extensive manual annotations and potentially generalizing to a broad span of tasks beyond the bias of the base annotated tasks. Such engineered pretext tasks for training a vision encoder include generative autoencoders \citep{autoencoders} or masked autoencoders \citep{MaskedAutoencoders2021}, multi-view contrastive learning \citep{chen2020simple}, or knowledge distillation using Teacher-Student models \citep{Caron2021EmergingPI}. Particulary for volumetric medical image analysis, UniMISS \citep{UniMiSS}, Model Genesis \citep{zhou2019models,zhou2021models}, and Swin UNETR default self-supervised weights \citep{SwinUNETRweights} correspond to this family and learning objectives. These strategies have been widely popular in recent years and have been practical solutions given the absence of large labeled datasets. Nevertheless, the collaborative endeavors of the medical imaging community to share open-access CT scans labeled at the voxel level have allowed us to gather a considerable amount of data with annotated anatomical structures (\textit{see} Table~\ref{datasets}). Thus, supervised pre-training, i.e., pre-training with the final objective task (segmentation), is emerging in medical imaging. For example, Med3D \citep{chen2019med3d}, MultiTalent \citep{multitalent}, CLIP-Driven model \citep{UniversalModel}, or SuPreM \citep{suprem} attest to this. There is a strong debate about which strategy is more suitable as a long-term solution since supervised pre-training requires high-quality annotations performed by clinical specialists, which is costly and time-consuming. Nevertheless, current empirical evidence \citep{suprem} suggests that supervised pre-training is more data-efficient than its unsupervised counterpart, even when transferred to novel tasks. 

In this work, we resort to supervised pre-training. We anticipate that our extensive experiments are aligned with the positive trends observed by concurrent studies \citep{UniversalModel, multitalent,suprem}, and are indeed exacerbated in the proposed data and parameter-efficient adaptation setting.
 
\subsection{Fine-tuning to downstream tasks}

Full fine-tuning \citep{autoencoders,finetuning} involves updating the weights of a pre-trained model using a supervised dataset corresponding to the target task. This has been the most straightforward and popular transfer learning approach in recent years, particularly for transferring knowledge across vision tasks \citep{Simonyan2014}. For the medical image analysis domain, this technique has become common practice in many applications, from radiology \citep{wang2017chestx} to retinal imaging \citep{de2018clinically,raghu2019transfusion}, usually using networks pre-trained on natural images. More recently, it has also become the favored strategy to adapt domain-specialized medical foundation models \citep{UniversalModel,multitalent}. Despite its popularity, fine-tuning has two main limitations. First, these methods are prone to over-fitting and catastrophic forgetting when the labeled data of the target task is insufficient \citep{partialFT,li2016learning,maml}. Second, this strategy may require substantial computing resources and lead to long adaptation times, especially when using large parameterized networks.

\subsection{Towards a more efficient adaptation of foundation models}

\subsubsection{Parameter-Efficient Fine-Tuning}
\label{ssec:rw_peft}

Aiming to transfer pre-trained deep representation models to multiple tasks by retaining its universal shareable features, PEFT was initially conceived in incremental learning \citep{blockmodular,Rusu2016ProgressiveNN}. The core idea is to avoid the destructive nature of fine-tuning on the previous knowledge acquired by such neural networks \citep{catforget,li2016learning}. Thus, the authors of these preliminary works proposed freezing the original network and incorporating new connections, so-called \textit{Adapters}, aiming to flexibly re-use old representations and learn new ones for each novel task. Such philosophy has come a long way in computer vision and, more concretely, in convolutional neural network (CNNs) architectures. Thus, several works developed PEFT strategies for transfer/incremental learning \citep{Rebuffi2017,Rebuffi2018}, or test-time adaptation \cite{wang2021tent} problems. More recently, with the increasing complexity of large neural networks, especially large language models based on Transformer architectures in natural language processing, such techniques have been revisited to exploit their parameter-efficient nature \citep{hu2022lora}. Fine-tuning these large foundation models requires unbearable computing resources, while tuning a small set of parameters alleviates such issues and hence allows the transfer of such networks to specialized tasks with commodity resources. 

From a technical standpoint, two main types of PEFT approaches exist: \textit{additive}, which incorporate a set of Adapters on the frozen encoder, and \textit{selective} methods, which only tune a subset of layers from the original model. Their particular variants may vary across the architecture employed, i.e., convolutional or Transformer blocks. Popular additive methods for CNNs include incorporating residual or sequential Adapters as stacked 1$\times$1 convolutions \citep{Rebuffi2017,Rebuffi2018}. For Transformer blocks, the recent and already very popular LoRA \citep{hu2022lora} embeds low-rank matrices in the self-attention layers of the Transformer model. Highly related, AdaptFormer \citep{adaptformer} embeds residual non-linear Adapters to modify the MLP feed-forward module. Note that low-rank Adapters such as LoRA are a special type of additive method, which performs a re-parametrization of the original weights. On the other hand, selective methods include tuning bias parameters on CNNs \citep{tinytl} or Transformers \citep{BenZaken2021BitFit}, or modifying the affine parameters in normalization layers, specifically batch \citep{frankle2021training} or layer \citep{layernorm}, particularly gamma and beta parameters, which act as a feature scaling filters.

Even though Parameter-Efficient Fine-Tuning has shown remarkable properties in different machine learning fields, its application in medical image analysis has shown limited adoption, with only a few works investigating image classification \citep{Dutt-missing-opportunity-23}. Thus, integrating PEFT on state-of-the-art medical volumetric segmentation networks remains unexplored.

\subsubsection{Black-box adaptation}

An important, practical consideration when choosing a transfer learning strategy is whether it operates in a black-box setting. For example, PEFT strategies assume knowledge of the pre-trained model architecture and are applied to specific layers. More importantly, they require explicit access to the model pre-trained weights, potentially resulting in source data leaking during adaptation \citep{leakdata}. These aspects may impede their deployment in privacy-preserving scenarios required in clinical scenarios. Indeed, in natural language processing and computer vision, there is currently an emerging literature on the fast few-shot adaptation using black-box models \citep{ColomboEMNLP2023,ouali2023black}, strongly motivated by the fact that large-scale foundation models are only available through online interfaces and their pre-trained weights are not shared. Therefore, model-agnostic Adapters that use only the features of the output layer of the pre-training model are highly relevant in practice and have yet to be explored in medical imaging. The primary baseline for black-box adaptation is linear probing, a linear logistic regression classifier typically used to measure the transferability capabilities of pre-trained models \citep{mahajan2018exploring}. More recently, Adapters have been introduced as small learnable modules that operate over feature representations. These include CLIP-Adapter, which introduces a non-linear MLP to project the original representations \citep{Gao2021}, memory-based methods \citep{Zhang2022}, or constraint learning objectives \citep{CLAP}.

Nevertheless, it is worth noting that not all these solutions directly apply to the scenario studied in this work, i.e., uni-modal, dense volumetric segmentation. For example, multi-modal solutions that largely depend on text supervision cannot be implemented in our scenario. At the same time, memory-based methods such as TIP-Adapter \citep{Zhang2022} would incur unbearable memory requirements for dense segmentation tasks. Therefore, proper methods for black-box adaptation for volumetric medical imaging segmentation have yet to be explored. In this work, we propose a novel modification of such Adapters to address this novel scenario.

%% Methods
%-----------

\begin{figure*}[ht!]
\centering
\includegraphics[width=.9\textwidth]{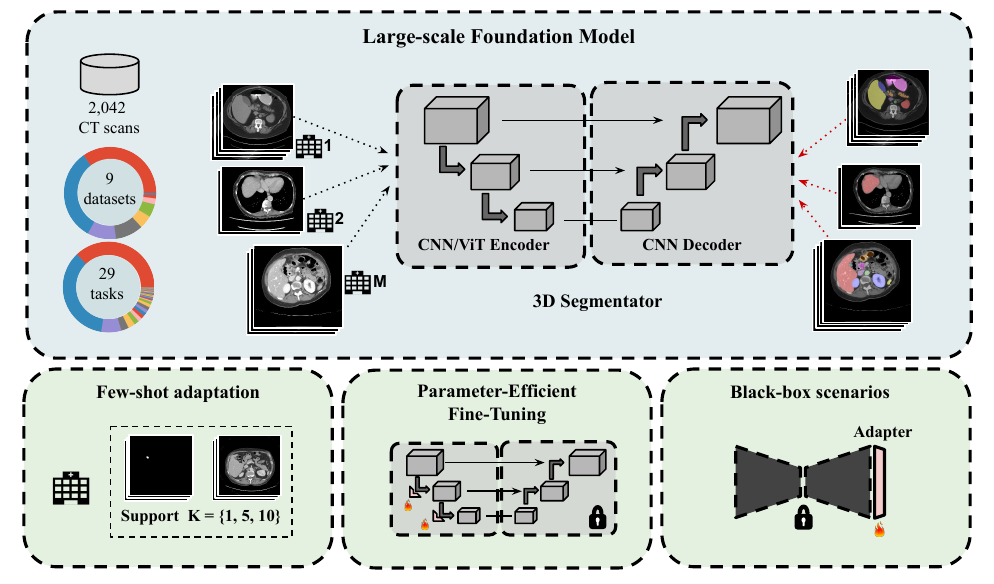}
\caption{\textbf{Few-Shot Efficient Fine-Tuning (FSEFT)}. A foundation model for volumetric medical images has been developed (\textit{see top}, and Section~\ref{sec:pretraining}). This model is pre-trained on a collection of CT volumes from nine open-access datasets, consisting of 2,042 scans with 29 partially labeled structures. Following Eq. \ref{eq:foundation_model}, the network is trained using a supervised learning objective, targeting organ segmentation. Then, a novel and realistic scenario is proposed, considering the resource limitations of clinical institutions for developing robust segmentation models. Concretely, this scenario assumes that: \textbf{\textit{(i) adaptation should be performed by accessing only a few labeled volumes}} (\textit{see bottom left} and Section~\ref{sec:fewshot}), and \textbf{\textit{(ii)}} the specialization of the foundation model to downstream tasks \textbf{\textit{should require the least computational resources}}. For the latter, Parameter-Efficient Fine-Tuning (PEFT) techniques such as LoRA, BitFit, or AdaptFormer (\textit{see bottom center} and Section~\ref{sec:peft}) and black-box Adapters such as linear probing and a newly proposed Spatial Adapter (\textit{see bottom right} and Section~\ref{sec:blackbox}), are considered.}
\label{fig:summary}
\end{figure*}

\section{Proposed Setting}
\label{sec:methods}

We introduce a pre-train-and-adapt scenario for medical volumetric foundation models that meets clinical real-world limitations during adaptation. In particular, transfer learning is performed in settings with limited data and computational resources. An overview of the framework is presented in Figure~\ref{fig:summary}. In the following, each of its components is formally introduced. 

\subsection{Foundation model training}
\label{sec:pretraining}

An assembly of $M$ different datasets, which integrate $N$ different volumes, is considered for training a large-scale foundation model. Let $\rmX_{n} \in \real^{\Omega_n}$ denotes a medical imaging volume, with $\Omega_n$ representing its spatial domain. Each volume is partially annotated at the voxel level, $\rmY_{n}=\{0, 1\}^{\Omega_n \times C}$, with $C$ the number of unique categories in the combined dataset. This means that some classes considered foreground in one dataset might be considered background in another set. Each dataset presents only partial categories annotated, known as a multi-label hot-encoding annotation vector. Thus, each image $\rmX_{n}$ is associated with the annotation vector corresponding to its dataset. To simplify the notation, this vector is denoted as $\rmw_{n}=\{0, 1\}^{C}$, and is directly associated with the dataset to which each sample belongs. Thus, the training set is composed of the input volumes, their corresponding partial labels, and annotation vectors: $\mathcal{D}_{T}=\{(\rmX_n, \rmY_n, \rmw_{n})\}_{n=1}^{N}$. Also, let us define a segmentation model, $\theta=\{\theta_f(\cdot), \theta_c(\cdot)\}$, which is composed of a feature extraction neural network, $\theta_f(\cdot)$, and a classification head, $\theta_c(\cdot)$. Thus, the backbone maps each voxel of the input into a spatial feature representation space, $\rmZ_n=\theta_f(\rmX_{n})$, with $\rmZ_n \in \real^{\Omega_n \times D}$ and $D$ the number of channels of the output features. Then, the classification head provides a probability distribution $\hat{\rmY}_{n} = \sigma(\theta_c(\rmZ_n))$, with $\sigma$ a sigmoid activation. Building the pre-trained foundation model $\theta$ amounts to using the curated assembled dataset by masked backpropagation of partial labels and optimizing any segmentation loss function, $\mathcal L_{SEG}$, using gradient descent: 
\begin{align}
\label{eq:foundation_model}
\min_{\theta_f,\theta_c} \quad & \frac{1}{\sum_c \rmw_{n,c}} \sum_c \rmw_{n,c} \mathcal L_{SEG}(\rmY_{n,c},\hat{\rmY}_{n,c})  , \quad n=1,...,N  \ \ .
\end{align}

\noindent Note that the masking operation for a given sample $n$, applied with $\rmw_{c}\in\{0, 1\}$, guides the contribution in weights update for such sample to consider only the segmentation loss for the specific categories labeled in its corresponding dataset.

\subsection{Few-Shot Efficient Fine-Tuning}
\label{subsec:method_fseft}

Formally, let us define a target dataset, $\mathcal{D}^*$, with volumes of an arbitrary study type, $X$, and a target organ to be segmented, $Y$. The goal is to adapt the pre-trained foundation model, $\theta$, to the new domain in an efficient way. A set of trainable parameters, $\phi$, are introduced for that purpose. These might be additional parameters to the foundation model or weights from the original base model, i.e., $\phi \subset \theta$, or a combination of both. These parameters are in charge of aligning the pre-trained features to potential domain drifts in the new dataset. 

In the following, we explore and propose novel alternatives to perform such adaptation requiring \textbf{\textit{(i)}} \textbf{\textit{low data resources}}, and \textbf{\textit{(ii)}} \textbf{\textit{being computationally efficient}}. Combined with specialized foundation models, such requirements consolidate the proposed \textbf{Few-Shot Efficient Fine-Tuning} framework for volumetric organ segmentation, a novel setting tailored to real-world clinical applications. We formalize a few-shot adaptation strategy to tackle the first constraint in Section~\ref{sec:fewshot}. For the latter requirement, we explore the use of recently popularized Parameter Efficient Fine-Tuning (PEFT) methods in Section~\ref{sec:peft} and propose novel black-box Adapters in Section~\ref{sec:blackbox}.

\subsubsection{Towards efficient adaptation using few-labeled volumes}
\label{sec:fewshot}

We assume that the adaptation should occur using only a few labeled examples, a.k.a. the \textit{support set} in the few-shot learning literature \citep{snell2017prototypical}, to alleviate the limitation in resources of the target institutions. Typically, a few-shot task includes a set of fully labeled support samples and one (or multiple) query samples. The support set is denoted as $\mathcal{D}_S=\{(X_{k}, Y_{k})\}_{k=1}^{\text{K}}$, with K the total number of support samples (so-called shots), which usually takes small values, i.e., $\text{K}=\{1,5,10\}$. The query (test) sample is denoted as a single volume $X$ for notation simplicity. The modified foundation model $\theta'=\{\theta, \phi\}$ yields sigmoid classification scores for both the query and support voxels: $\forall x \in X, \hat{Y}(x) = \sigma(\theta'(x))$ and $\forall x \in X_k, \hat{Y}_k(x) = \sigma(\theta'(x)), k \in 1, \dots \text{K}$. Thus, adapting the foundation model in this setting involves optimizing the set of tuning parameters $\phi$ to the new task by minimizing a target segmentation loss function on the few support samples:
 \begin{align}
\label{eq:general_adaptation}
\min_{\phi} \quad & \sum \mathcal L_{SEG}(\rmY_{k},\hat{\rmY}_{k}) , \quad k=1,...,\text{K}  \ \ ,
\end{align}

\subsubsection{Parameter-Efficient Fine-Tuning}
\label{sec:peft}

A commonly used strategy for adapting a pre-trained model to new tasks is updating the whole model ($\phi=\theta$), a.k.a. fine-tuning. Nevertheless, this strategy is largely computationally expensive and might show poor performance when available training data is scarce (\textit{see} Figure~\ref{fig:intro}(a) and (b)). A relevant core of popular prior literature, the so-called Parameter-Efficient Fine-Tuning, studies the update of a smaller set of parameters instead. Different applications in natural language processing or general computer vision communities have demonstrated resource-efficient adaptation on large-parameterized networks by notably reducing peak memory usage. Such solutions are an appealing direction to tackle the proposed FSEFT setting, thus meeting the computational limitations of clinical institutions. In addition, constraining the learning process to perform minor modifications on the learned feature representations might avoid catastrophic forgetting. Thus, we propose integrating popular PEFT methods (\textit{see} Section~\ref{ssec:rw_peft}) as an alternative to tackle the described FSEFT setting.

\subsubsection{Black-box adaptation}
\label{sec:blackbox}

An appealing direction for privacy-preserving scenarios is using solely the output pre-trained features produced by the foundation model, i.e., $\rmZ$ in our notation. These strategies, so-called black-box Adapters \citep{ouali2023black}, do not require access to model weights, thus avoiding potential proprietary and private data leaking \citep{leakdata}. Linear probe and Adapter-style methods are popularly considered in this setting. Building upon such methods, a novel Spatial Adapter is proposed, tailored to dense prediction tasks, as described below.

\textit{\textbf{Linear probe}}. It consists of only tuning a linear classification head. The adjusted parameters are $\phi_c = \{\mathcal{W}^{D\times 1}, \mathcal{B}^1\}$, being $\mathcal{W}$ the target class prototype, and $\mathcal{B}$ the bias. Nevertheless, such a baseline method, typically employed to assess the transferability capabilities of pre-trained feature representations, neglects non-linear feature relationships on the target task.

\textit{\textbf{Adapter-style}}. To tackle this limitation, CLIP-Adapter \citep{Gao2021} introduced, in the context of vision-language models, an Adapter that refines the pre-trained features by adding a set of non-linear operations, $\phi_{f}$. These layers perform a non-linear low-rank modification that refines the original feature projection, such that $\rmZ' = \rmZ + \phi_{f}(\rmZ)$. Afterward, such modified feature representation is forwarded through a linear classifier to output predicted scores, such that $\hat{Y} = \phi_c(\rmZ')$. Nevertheless, as we later demonstrate empirically (\textit{see} Section~\ref{ssec:ablation}), this MLP-based Adapter might fail to capture rich local information, which is natural for dense segmentation problems. To alleviate this issue, we propose Spatial Adapters, in which $\phi_{f}$ comprises a stacked combination of 3D convolutional filters, a black-box strategy specially designed for volumetric medical image segmentation.

\subsection{Leveraging anatomical priors during adaptation}
\label{subsec:method_ti}

\begin{figure}[h!]
\centering
\includegraphics[width=.48\textwidth]{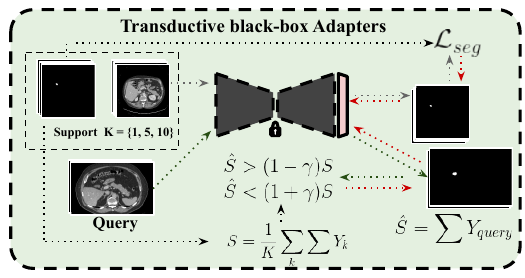}
\caption{\textbf{Leveraging anatomical priors}. We propose integrating priors on the target structure size while adapting the pre-trained foundation model in a transductive fashion. Concretely, a black-box Adapter is trained to minimize a given segmentation loss on the few support annotated volumes, and an inequality constraint is applied during training to produce consistent organ sizes (\textit{see} Eq. \ref{eq:transductive_inference}).}
\label{fig:summary_transductive}
\end{figure}

In image classification, inference is often performed in an \textit{inductive} manner, i.e., one sample at a time. This inductive inference paradigm is common in medical image segmentation, where the task is often seen as a voxel classification. However, segmentation is a \textit{transductive} problem by nature, i.e., one could make joint predictions for all the voxels of the test subject, leveraging available priors on the global structure of the output mask, such as the shape of the target organ. Thus, transduction is appealing in our few-shot medical image segmentation setting as the support set could provide such approximate priors. Specifically, we propose to perform transductive inference, optimizing the segmentation loss on the support set while imposing inequality constraints on the size of the target organ in the test subject. Figure~\ref{fig:summary_transductive} presents an overview of the proposed setting, which is now formalized. 

\textit{\textbf{Proportion constraints on the query sample}}. Since the volumes are pre-processed to a homogeneous resolution, one could estimate an average target-region proportion from the support samples as follows: $S=\frac{1}{K}\sum_k \sum_{x \in \Omega} Y_k (x)$, with $\Omega$ denoting the 3D spatial domain. Now, let $\hat{S}$ denote the predicted size of the target region in the test image as the summation of sigmoid output over the spatial image domain: $\hat{S}=\sum_{x \in \Omega} \hat{Y}(x)$. Thus, the following term is incorporated during inference using black-box Adapters, penalizing region proportions that differ from the target by a margin $m$ from the anatomical priors: 
\begin{align}
\label{eq:size_penalty}
       \mathcal{L}_{TI}=
       \begin{cases} 
        |\hat{S} - (1-m)S|, & \text{if}\ \hat{S} < (1-m)S\\
        |\hat{S} - (1+m)S|, & \text{if}\ \hat{S} > (1+m)S\\
        0, & \text{otherwise}
        \end{cases} \ \ ,
\end{align}

\noindent where the margin $m$ serves as a relaxation of the constraint to account for the anatomical variability and potential noise in estimating the target organ size in the few support shots. Thus, divergences smaller than $m$ between estimated and target sizes will not result in effective penalties. Such value is empirically set to all tasks the same (\textit{see} Section~\ref{sec:bb_implementation}), as its effect is further explored in respective ablation studies in Section~\ref{ssec:ablation}.

\textit{\textbf{Combined learning objective}}. The black-box Adapter is trained by gradient steps, integrating the segmentation loss on the support samples and the transductive anatomical constraints in Eq. \ref{eq:size_penalty} to the estimated organ size of the query sample, $\hat{S}_{query}$. The overall learning objective is consolidated as follows:
 \begin{align}
\label{eq:transductive_inference}
\min_{\phi_f,\phi_c} \quad &  \mathcal L_{SEG}(Y_{k},\hat{Y}_{k}) + \lambda \mathcal L_{TI}(S, \hat{S}_{query})  , \quad k=1,...,\text{K} \ \ ,  
\end{align}

\noindent where $\lambda$ is a blending hyper-parameter that controls the relative importance of both terms, empirically fixed in Section~\ref{sec:bb_implementation}.

%% Experiments and results
%-----------
\section{Experiments}
\label{sec:experiments}

\subsection{Datasets}

\subsubsection{Assemblying datasets}

Publicly available datasets of CT volumes are used to build the foundation model and perform the adaptation experiments. Table~\ref{datasets} presents a summary of these datasets.

\begin{table*}[h!]
\centering
\caption{\textbf{Assembly of CT whole-body volumes datasets from various open-access sources.} A foundation model for volumetric organ segmentation in CT scans is proposed, compiling nine publicly available datasets, 29 different structures annotated at the voxel level, and 2,042 volumetric samples used for supervised pre-training. In addition, the efficient adaptation capabilities of the foundation model are evaluated in two external sources, 25 tasks of which have been included.}
\label{datasets}
\scriptsize{
\begin{tabular}{lrrl}
\toprule
\multicolumn{1}{l}{Dataset} & \multicolumn{1}{c}{\#Categories} & \multicolumn{1}{c}{\#Scans} & \multicolumn{1}{l}{Selected tasks} \\
\midrule
\multicolumn{4}{l}{\textbf{Foundation model training}} \\ \hdashline\noalign{\vskip 0.5ex}
BTCV \citep{BTCV}                     & 13          & 25      & Spl, rKid, lKid, Gall, Eso, Liv, Sto, Aor, PostC, PSV, Pan, rAd, lAd.                 \\
CHAOS \citep{CHAOS}                   & 1           & 14      & Liv.                                                                                  \\
LiTS \citep{LiTS}                     & 2           & 95      & Liv, LivTum.                                                                          \\
KiTS \citep{KiTS}                     & 5           & 204     & rKid, KidTum, KidCyst, lKid.                                                          \\
AbdomenCT-1K \citep{Abdomen-CT}       & 5           & 706     & Liv. rKid, Spl, Pan, lKid.                                                            \\
AMOS \citep{AMOS}                     & 15          & 200     & Spl, rKid, lKid, Gall, Eso, Liv, Sto, Aor, PostC, Pan, rAd, lAd, Duo, Blad, pr/ut.    \\
MSD subtasks \citep{MSD}              & 9           & 661     & Liv, LivTum, LungTum, Pan, PanTum, hv, hvt, Spl, ColonTum.                            \\
AbdomenCT-12organ \citep{Abdomen-CT}  & 13          & 94      & Liv, rKid, SPl, Pan, Aor, PostC, Sto, Gall, Eso, rAd, lAd, celTruck, lKid.            \\
CT-ORG \citep{CT-ORG}                 & 8           & 43      & Liv, Bladd, rLung, rKid, Bone, Brain, lLung, lKid.                                    \\ 
\our \textbf{Total}                                      & \our \textbf{29}          & \our \textbf{2,042}    & \our                                                                                      \\
\midrule
\multicolumn{4}{l}{\textbf{Few-Shot Efficient Fine-Tuning}}                                                                                                              \\ \hdashline\noalign{\vskip 0.5ex}
TotalSegmentator \citep{TotalSeg}     & 25           & 1,203   & \textit{Base}: Spl, lKid, Gall, Eso, Liv, Pan, Sto, Duo, Aor.                                          \\
                                      &            &    & \textit{Novel}: Lung, heart, and gluteus parcellation.                                        \\
FLARE'22 \citep{FLARE22}       & 9           & 72    & Spl, lKid, Gall, Eso, Liv, Pan, Sto, Duo, Aor.                                          \\
\bottomrule
\multicolumn{4}{l}{\notsotiny {Abbreviations:  Spl: spleen; rKid: kidney right; lKid: kidney left; Gall: gallbladder, Eso: esophagus; Liv: liver; Sto: stomach; Aor: aorta; PostC: postcava vein; PSV: portal and splenic vein;}} \\
\multicolumn{4}{l}{\notsotiny {Pan: pancreas; rAd: right adrenal gland; lAd: left adrenal gland; Duo: duodenum; Blad: bladder; pr/ut: prostate/uterus; LivTum: liver tumor; KidTum: kidney tumor; KidCyst: kidney cyst;}} \\
\multicolumn{4}{l}{\notsotiny {celTruck: celiac truck; rLung: right lung; lLund: left lung; LungTum: lung tumor; PanTum: pancreas tumor; hv: hepatic vessel; hvt: hepatic tumor; ColonTum: colon tumor.}} \\
\end{tabular}
}
\end{table*}

\textit{\textbf{Foundation model training}}. A total of nine datasets with $29$ different anatomical structures annotated are assembled. Concretely, BTCV \citep{BTCV}, CHAOS \citep{CHAOS}, LiTS \citep{LiTS}, KiTS \citep{KiTS}, AbdomenCT-1K \citep{Abdomen-CT}, AMOS \citep{AMOS}, MSD subtasks \citep{MSD}, AbdomenCT-12organ \citep{Abdomen-CT} and CT-ORG \citep{CT-ORG} are gathered to retrieve up to $2,042$ CT volumes for training. These datasets include several annotated structures at the voxel level, including organs (e.g., spleen liver, esophagus, kidneys, etc.) and lesions (e.g., liver tumor, kidney cyst, etc.) from full thorax CT scans. Each dataset has partially labeled scans, meaning only a few categories are labeled for each data source.

\textit{\textbf{Transferability}}. TotalSegmentator \citep{TotalSeg} and FLARE'22 \citep{FLARE22} are used to evaluate the adaptation of the foundation model to external domains. TotalSegmentator comprises 1,024 CT volumes with up to 104 anatomical structures and a wide heterogeneity of scanners and study types. FLARE'22 \citep{FLARE22} is a large dataset containing fully annotated CT scans and unlabeled samples. In particular, the training subset consists of 50 scans, for which 13 organs are manually segmented at the voxel level. It is worth mentioning that FLARE'22 contains CT scans of patients with liver, kidney, spleen, or pancreas diseases, which poses a challenging domain shift with respect to the pre-training data. In addition, both data sources represent different demographics, thus ensuring a robust assessment of the foundation model performance. First, the efficient adaptation to \textbf{\textit{(i)} \textit{base categories}} is explored. Since the number of possible structures in the human body is a narrow set of fixed categories, a realistic scenario is considered in which the foundation model has been trained to segment the target organ, but domain drifts exist in the institution, acquisition device, or protocol. Retrieved cases corresponding to nine representative organs found in the foundation model training data are selected, i.e., spleen, left kidney, gallbladder, esophagus, liver, pancreas, stomach, duodenum, and aorta. Experiments in TotalSegmentator are carried out in a binary classification fashion, i.e., one organ at a time, to simulate a real-world use case during adaptation. Additionally, FLARE'22 is exploited to investigate multi-class segmentation of the same structures. Second, the transferability of the proposed supervised pre-trained model to \textbf{\textit{(ii)} \textit{novel tasks}} is assessed. Such experiments are carried out on the TotalSegmentator dataset, addressing the lung, heart, and gluteus parcellation. Note that these tasks involve segmenting 16 different anatomical structures.

\subsubsection{Pre-processing and standardization}

Following previous works \citep{UniversalModel}, all volumes are standardized and pre-processed to reduce the inter-domain gap. In particular, the orientation of CT volumes is fixed, and isotropic spacing is used to resample the volume voxels to cubes of size $1.5 \ mm^3$. Finally, the intensity range is clipped to the range [-175, 250] and linearly scaled to [0, 1].

\subsection{Foundation model training}

\subsubsection{Model architecture}

Swin-UNETR \citep{hatamizadeh2022swin} is employed as the model architecture for volumetric organ segmentation. Swin-UNETR is a hybrid U-Net architecture, composed of a Swin Transformer encoder \citep{swinTransf}, and a CNN-based decoder composed of residual blocks. In particular, the employed Swin-UNETR architecture width is designed to output a representation of 48 features per voxel. The encoder weights were initialized using the self-supervised pre-pretrained representations in \citep{SwinUNETRweights}.

\subsubsection{Training details}

The proposed foundation model is trained on the 29 tasks defined in the assembly dataset (\textit{see} Table~\ref{datasets}), by optimizing the Dice loss in Eq. \ref{eq:foundation_model}. In each step, three input patches of size $96\times96\times96$ per volume are forwarded to the model, using a batch size of two volumes during $120$ epochs and four distributed A6000 GPUs. The optimizer employed is AdamW. The base learning rate is set to $10^{-4}$, and a warm-up cosine scheduler of ten epochs is included. Input patches are augmented through intensity shifts and random rotations of $90$ degrees.

\subsection{Adapting foundation models to downstream datasets}
\label{subsec:adapting_fm}

Once trained, the efficient transferability capabilities of the proposed foundation model are assessed by adapting it to downstream domains or tasks. Following the FSEFT setting detailed in Section~\ref{subsec:method_fseft}, data and parameter efficiency are considered the main goals of such a process. The experimental setting details and implementation of popular PEFT methods, black-box Adapters, and baselines are described below.

\subsubsection{Adaptation resources}

Two different data regimes are considered within the data-efficient FSEFT setting formalized in Section~\ref{sec:fewshot}: \textbf{\textit{(i)} \textit{low data regime}}, in which K=$\{1, 5, 10\}$ annotated volumes are used, and \textbf{\textit{(ii)} \textit{large data regime}} where a more significant number of scans, i.e., K=30, are considered. It is worth highlighting that the proposed large data regime is significantly more strict than the transfer scenarios considered in concurrent literature \citep{UniversalModel,multitalent}, where the authors profit up to hundreds of samples for adaptation.

\subsubsection{General implementation details for adaptation}
\label{sec:train_initialization}

\textit{\textbf{Classification head design}}. Following relevant literature in transfer learning from supervised pre-trained models for natural image understanding \citep{mlpproj}, the projection (if existing) and classification heads are removed from the foundation model, and a new linear layer is incorporated to tackle the novel set of tasks. Then, standard literature commonly initializes a new classification head using random weights. In contrast, we propose using the pre-trained class prototypes and bias parameters for each structure learned by the foundation model for initialization instead, should such an organ be included during the pre-training stage. For the binary segmentation scenario, i.e., TotalSegmentator, sigmoid activation is used, whereas in the case of FLARE'22, the multi-class segmentation scenario using softmax outputs is explored.

\textit{\textbf{Training details}}. The foundation model is adapted using the Dice loss as the target learning objective. Furthermore, a strict few-shot setting is explored, in which no validation data is available during adaptation. AdamW is used as an optimizer, and a cosine decay learning rate scheduler of 200 epochs is defined for training. Since each adaptation method might present a particular optimum learning rate, the initial value for each adaptation strategy is set individually. Therefore, each value is adjusted empirically, based on convergence on the support set within 20 epochs in a grid-search using the spleen, left kidney, and gallbladder organs of TotalSegmentator. Afterward, the learning rate is fixed across all tasks and data regimes. In addition, the training loss is monitored in the support set, and early stopping is applied upon convergence, assuming that its value does not increase more than $0.001$ during 20 epochs. The model takes as input six patches of size $96\times96\times96$ per volume in each iteration with a batch size of one volume. Intensity shifts are applied for data augmentation across all the methods. These experiments are carried out using a single A6000 GPU. 

\subsubsection{Parameter-Efficient Fine-Tuning}
\label{sec:peft_implementation}

As introduced in Section~\ref{sec:peft}, PEFT is an appealing solution for the proposed FSEFT setting. Popular segmentation backbones for medical segmentation build upon encoder-decoder architectures. In this work, we advocate modifying the foundation model encoder, which propagates the effect throughout the entire model. 

\textit{\textbf{Encoder tuning}}. Regarding Transformer-based encoders, e.g., Swin-UNETR \citep{hatamizadeh2022swin}, the following strong baselines are considered: BitFit \citep{BenZaken2021BitFit}, AdaptFormer \citep{adaptformer}, LoRA \citep{hu2022lora}, and adapting Affine parameters in normalization layers \citep{frankle2021training} (Affine-LN). For purely convolutional architectures, e.g., U-Net \citep{unet}, analogously, Bias \citep{tinytl} and Affine parameters from batch normalization layers \citep{frankle2021training} (Affine-BN) tuning are included. Additionally, the residual Adapter proposed in \citep{Rebuffi2018} for CNNs, inserting a set of 1$\times$1 convolutional kernels residually into encoder blocks, is employed.

\textit{\textbf{Decoder transferability}}. For \textit{base tasks}, the rich reconstructions learned during pre-training are exploited during transfer learning, and thus the decoder is frozen. This allows for an extremely parameter-efficient adaptation. Regarding \textit{novel organs}, the decoder is tuned in combination with the PEFT of the encoder representations.

\textit{\textbf{Implementation details}}. Next, each approach's specific tuned parameters and hyper-parameters are listed individually. For BitFit, the bias parameters of both the attention and MLP layers in the Transformer blocks are tuned, with an initial learning rate of $10^{-2}$. Regarding LoRA, low-rank matrices with a rank value of $r=4$ are incorporated into the key-query-value layers of each Transformer block. The base learning rate for LoRA is set to $10^{-3}$. Analogously, for AdaptFormer, the re-parametrization rank is set to 16, and the residual Adapter is added to the Transformer MLP layer. In this case, the best learning rate found for convergence was $10^{-2}$. Finally, the feature scaling parameters $\beta$ and $\gamma$ of the instance/batch normalization layers in the encoder are updated in Affine-LN/BN, using a learning rate of $10^{-2}$. When combining PEFT methods with decoder fine-tuning, the base learning rate is set to $10^{-4}$.

\subsubsection{Black-box Adapters}
\label{sec:bb_implementation}

Appending black-box Adapters over the pre-trained features, i.e., the model is completely frozen during adaptation, is proposed as an efficient privacy-preserving alternative using linear probes and the proposed Spatial Adapters.

\textit{\textbf{Implementation details}}. For linear probing, only the classification head is updated via gradient descent. A base learning rate of $10^{-2}$ is fixed. Second, the proposed Spatial Adapter is used to modify the pre-trained features, accounting for the 3D patterns in the feature space. Concretely, two convolutional blocks with a kernel size of three and a constant number of channels are residually applied, using a learning rate of $10^{-4}$. 

\textit{\textbf{Transductive inference}}. Finally, black-box Adapters are enhanced by integrating priors on the target organ size, as described in Section~\ref{subsec:method_ti}. In this scenario, the size constraint in Eq.~\ref{eq:size_penalty} is applied over the prediction of the query sample, by setting $\lambda=1$, and $m=0.1$.

\subsubsection{Baselines}

\textit{\textbf{Zero-shot}}. First, the direct inference of the foundation model on the target domain, commonly referred to as zero-shot predictions in the literature, i.e., no adaptation is carried out, is used as a lower performance bound for the known tasks. 

\textit{\textbf{Full fine-tuning}}. Second, standard full fine-tuning is considered to evaluate the potential of the proposed setting. Fine-tuning from self-supervised pre-training is considered, using the model weights in \cite{SwinUNETRweights}. By exploiting 5,050 CT scans, they represent the current state-of-the-art for unsupervised volumetric pre-training. The Swin-UNETR encoder is initialized using these weights, and a new decoder is randomly initialized. Then, the whole model is fine-tuned in the few-shot data regime. Analogously, full fine-tuning of our pre-trained supervised foundation model (\textit{Ours}) is carried out. For these strategies, the base learning rate is decreased to $10^{-4}$.

\subsection{Evaluation protocol}

The different methods are evaluated on an independent subset of query samples from each dataset. The adaptation-evaluation process is repeated using three random seeds to account for the variability inherent to the few-shot setting. During testing, the predicted sigmoid probabilities are thresholded by a fixed value of $0.5$ for the binary scenario. In the case of multi-class segmentation tasks, the category with the maximum probability is selected for each voxel. Last, the Dice similarity coefficient (DSC) per organ is extracted as the figure of merit.

\section{Results}
\label{sec:results}

\subsection{Transferability on TotalSegmentator}

\subsubsection{Few-shot adaptation performance}

Results obtained adapting our pre-trained foundation model on TotalSegmentator base tasks are reported in Table~\ref{results_ours}. A graphical representation of such results is also introduced in Figure~\ref{fig:intro}(a). In the following, the main findings are detailed.

\begin{table*}[h!]
\centering
\caption{\textbf{Few-shot efficient adaptation results using the proposed foundation model (\textit{Ours}) on TotalSegmentator}. Adaptation is performed for each organ individually in a binary segmentation task. The metric presented is Dice. The best results are highlighted in bold, and the second-best performance is underscored. FULL: Whole model training; PEFT: Parameter-Efficient Fine-Tuning; BB: black-box. The proposed black-box Adapter performance is shadowed.}
\label{results_ours}
\scriptsize{
\begin{tabular}{cllcccccccccc}
\toprule
\multicolumn{2}{c}{Setting} & \multicolumn{1}{c}{Method}                                                   & Spl   & lKid  & Gall  & Eso   & Liv   & Pan   & Sto   & Duo   & Aor   & \textbf{Avg.}  \\ \midrule
&                                                          & Zero-shot                                     & 91.34 & 89.05 & 77.18 & 37.72 & 93.03 & 78.15 & 75.86 & 44.00 & 66.35 & 72.41 \\
\midrule 
\multirow{8}{*}{1-shot} & \multirow{2}{*}{FULL}           & Fine-tuning \citep{SwinUNETRweights}           &  \multicolumn{1}{r}{5.00} &  \multicolumn{1}{r}{2.51} &  \multicolumn{1}{r}{0.67} &  \multicolumn{1}{r}{1.35} & 22.27 &  \multicolumn{1}{r}{0.53} &  \multicolumn{1}{r}{2.02} &  \multicolumn{1}{r}{0.29} &  \multicolumn{1}{r}{4.83} &  \multicolumn{1}{r}{4.39} \\ 
&                                                         & Fine-tuning (\textit{Ours})                    & 91.31 & 88.00 & 76.58 & 49.11 & 98.80 & 77.31 & 68.54 & 58.24 & 81.17 & \underline{75.90}\\ \cdashlinelr{2-13}
& \multirow{4}{*}{PEFT}                                   & BitFit \citep{BenZaken2021BitFit}              & 88.42 & 87.89 & 75.49 & 48.23 & 92.00 & 74.92 & 70.27 & 49.95 & 72.03 & 73.24 \\
&                                                         & LoRA \citep{hu2022lora}                        & 90.38 & 88.68 & 77.63 & 44.54 & 92.48 & 78.13 & 72.19 & 54.75 & 70.28 & 74.34 \\
&                                                         & AdaptFormer \citep{adaptformer}                & 89.85 & 88.75 & 81.22 & 47.64 & 89.95 & 68.97 & 70.63 & 57.45 & 76.47 & 74.55 \\
&                                                         & Affine-LN \citep{layernorm}                    & 91.62 & 88.87 & 79.34 & 45.57 & 90.17 & 75.05 & 67.43 & 49.76 & 71.78 & 73.29 \\ \cdashlinelr{2-13}
& \multirow{2}{*}{BB}                                     & Linear probe                                   & 91.33 & 88.74 & 78.22 & 45.27 & 92.94 & 78.23 & 76.10 & 57.63 & 67.08 & 75.06 \\
&                                                         & \our Spatial Adapter                           & \our91.75 & \our88.74 & \our77.92 & \our47.01 & \our92.77 & \our77.41 & \our75.89 & \our65.82 & \our68.94 & \our \textbf{76.25} \\
\midrule 
\multirow{8}{*}{5-shot} & \multirow{2}{*}{FULL}            & Fine-tuning \citep{SwinUNETRweights}           & 19.38 & 13.29 & 13.22 & 45.48 & 34.63 & 34.87 & 17.06 & \multicolumn{1}{r}{4.02}  & 44.72 & 25.19 \\
&                                                          & Fine-tuning (\textit{Ours})                    & 90.47 & 89.02 & 71.88 & 54.33 & 93.75 & 78.20 & 59.26 & 66.27 & 91.20 & \textbf{77.15}  \\ \cdashlinelr{2-13}
& \multirow{4}{*}{PEFT}                                    & BitFit \citep{BenZaken2021BitFit}              & 91.46 & 88.26 & 71.37 & 50.50 & 92.96 & 57.94 & 71.22 & 56.91 & 73.65 & 72.70  \\
&                                                          & LoRA \citep{hu2022lora}                        & 92.18 & 89.08 & 73.69 & 48.98 & 93.18 & 78.27 & 72.73 & 60.18 & 78.43 & 76.30  \\
&                                                          & AdaptFormer \citep{adaptformer}                & 90.32 & 89.49 & 74.93 & 45.54 & 93.16 & 71.36 & 74.01 & 64.62 & 83.22 & 76.29  \\
&                                                          & Affine-LN \citep{layernorm}                    & 91.75 & 90.19 & 76.86 & 47.51 & 93.30 & 77.25 & 74.15 & 63.56 & 74.66 & 76.58 \\ \cdashlinelr{2-13}
& \multirow{2}{*}{BB}                                      & Linear probe                                   & 91.40 & 89.51 & 78.14 & 45.98 & 92.68 & 78.31 & 76.64 & 63.27 & 69.06 & 76.11 \\
&                                                          & \our Spatial Adapter                           & \our91.97 & \our90.03 & \our76.82 & \our46.36 & \our92.95 & \our78.85 & \our77.03 & \our66.41 & \our70.58 & \our \underline{76.78} \\
\midrule
\multirow{8}{*}{10-shot} & \multirow{2}{*}{FULL}           & Fine-tuning \citep{SwinUNETRweights}           & 33.27 & 36.14 & 24.55 & 51.90 & 62.86 & 54.24 & 31.00 & 22.87 & 76.87 & 43.74 \\
&                                                          & Fine-tuning (\textit{Ours})                    & 92.97 & 80.85 & 72.96 & 56.56 & 92.36 & 78.67 & 61.27 & 66.56 & 90.15 & 76.93 \\ \cdashlinelr{2-13}
& \multirow{4}{*}{PEFT}                                    & BitFit \citep{BenZaken2021BitFit}              & 92.51 & 89.41 & 76.48 & 50.57 & 92.54 & 79.40 & 67.09 & 65.32 & 76.89 & 76.69 \\
&                                                          & LoRA \citep{hu2022lora}                        & 91.91 & 90.10 & 82.66 & 49.36 & 93.45 & 80.86 & 67.97 & 61.36 & 82.47 & \textbf{77.79} \\
&                                                          & AdaptFormer \citep{adaptformer}                & 90.36 & 89.91 & 73.67 & 49.19 & 89.50 & 71.13 & 66.85 & 64.95 & 80.48 & 75.12 \\
&                                                          & Affine-LN \citep{layernorm}                    & 92.20 & 86.02 & 79.58 & 50.27 & 89.98 & 77.64 & 69.15 & 67.64 & 83.53 & 77.33 \\ \cdashlinelr{2-13}
& \multirow{2}{*}{BB}                                      & Linear probe                                   & 91.72 & 89.78 & 78.48 & 46.48 & 92.16 & 78.14 & 76.81 & 63.63 & 69.92 & 76.35 \\
&                                                          & \our Spatial Adapter                           & \our92.25 & \our89.86 & \our78.45 & \our49.77 & \our92.08 & \our78.79 & \our76.97 & \our66.73 & \our71.27 & \our \underline{77.35} \\
\midrule 
\multirow{8}{*}{30-shot} & \multirow{2}{*}{FULL}           & Fine-tuning \citep{SwinUNETRweights}           & 36.56 & 66.78 & 61.25 & 63.65 & 88.99 & 72.92 & 56.26 & 68.22 & 94.44 & 67.67 \\
&                                                          & Fine-tuning (\textit{Ours})                    & 94.47 & 90.75 & 67.06 & 53.71 & 92.36 & 80.65 & 64.08 & 79.38 & 92.55 & \underline{79.45} \\ \cdashlinelr{2-13}
& \multirow{4}{*}{PEFT}                                    & BitFit \citep{BenZaken2021BitFit}              & 94.35 & 89.57 & 80.34 & 51.10 & 92.83 & 82.17 & 65.93 & 65.89 & 83.64 & 78.42 \\
&                                                          & LoRA \citep{hu2022lora}                        & 94.01 & 89.01 & 84.98 & 53.95 & 94.07 & 81.75 & 69.68 & 65.79 & 87.12 & \textbf{80.04} \\
&                                                          & AdaptFormer \citep{adaptformer}                & 92.62 & 85.28 & 83.98 & 42.96 & 90.37 & 78.70 & 71.33 & 70.01 & 87.26 & 78.06 \\
&                                                          & Affine-LN \citep{layernorm}                    & 94.92 & 86.00 & 75.91 & 53.38 & 92.97 & 81.50 & 67.72 & 69.12 & 88.05 & 78.84 \\ \cdashlinelr{2-13} 
& \multirow{2}{*}{BB}                                      & Linear probe                                   & 92.92 & 90.52 & 81.47 & 46.46 & 92.53 & 78.35 & 77.06 & 64.64 & 70.70 & 77.18 \\
&                                                          & \our Spatial Adapter                           & \our93.19 & \our89.64 & \our81.16 & \our49.85 & \our92.55 & \our79.76 & \our78.45 & \our69.50 & \our71.88 & \our78.44 \\
\bottomrule
\end{tabular}
}
\end{table*}

\textit{\textbf{Large data regime}}. Results with K=30 support samples (Table~\ref{results_ours}, \textit{bottom}) show that the proposed supervised pre-training brings substantial improvements compared to fine-tuning from self-supervised pre-trained models, with impressive average gains of $+11.8\%$ compared to the state-of-the-art weights from \cite{SwinUNETRweights}. LoRA is the best-performing parameter-efficient adaptation strategy, outperforming full fine-tuning. Despite black-box Adapters yielding slightly lower metrics, they outperform full fine-tuning from self-supervised weights and perform similarly to competitive PEFT approaches, e.g., BitFit and Affine-LN. When the proposed Spatial Adapter is integrated, the overall performance increases by $+1.3\%$, surpassing PEFT methods such as AdaptFormer. It is worth recalling that black-box Adapters can overcome privacy issues since they potentially do not require access to input data or model weights during adaptation. In contrast, they directly operate over pre-trained features, contrary to PEFT or full training strategies. Hence, their promising performance is of special interest in medical applications.

\textit{\textbf{Low data regime}}. As well-established knowledge in the deep learning community, full training strategies considerably decrease their performance in low-labeled data regimes, which is supported by the results in Table~\ref{results_ours} (e.g., K=1). Note that this observation is aligned with concurrent works fine-tuning pre-trained models on TotalSegmentator \citep{suprem}. PEFT methods might alleviate this issue, as they better retain the pre-trained knowledge, which might mitigate severe over-fitting to the labeled data points used for adaptation (e.g., K=10). Nevertheless, they still show worse capabilities than black-box Adapters for known tasks. For example, the best-performing method of this family when K=5, i.e., Affine-LN, falls behind Spatial Adapters by $+0.2\%$. More interestingly, black-box strategies show strong robustness in extreme settings where only one annotated volume is available during adaptation. As a figure regarding the potential of the proposed supervised pre-training, the obtained results using only one annotated volume are $+8.6\%$ better than fine-tuning the whole model from self-supervised pre-training using 30 annotated volumes. Including the proposed Spatial Adapter module to refine the pre-trained features brings promising improvements over linear probing across all number of shots. These gains are especially relevant in the target organs where the foundation model shows worse direct generalization, such as the esophagus or duodenum.

\subsubsection{Leveraging anatomical priors}

In the previous section, the use of the proposed Adapters has been empirically motivated. In addition to these adapters, the proposed formulation in Section~\ref{subsec:method_ti} integrates a constrained transductive inference. This inference aims to exploit task-specific knowledge. More concretely, given some priors on the target task, such as the expected approximate volume in organ segmentation, the transductive inference can leverage this information via constraints on the query sample. 

\textit{\textbf{Performance analysis}}. The obtained results when adapting the pre-trained foundation model in the low-labeled data regime using black-box strategies, i.e., linear probing and a Spatial Adapter, are presented in Table~\ref{transductive_table}. Results show consistent improvements ranging from $+0.4\%$ to $+0.9\%$ for both linear probing and the proposed Adapter. Concretely, the largest impact is obtained when K=1, which suggests that incorporating anatomical priors on the target task might provide larger performance gains when less support data is accessible for adaptation. Interestingly, tuning the proposed Adapter in this transductive setting when using 10 support volumes approaches PEFT methods, trained on the large data regime, i.e., using K=30 shots (\textit{see} Table~\ref{results_ours}), and reduces the gap with full fine-tuning, while requiring much fewer samples. 

\begin{figure}[t!]
\centering
\includegraphics[width=.43\textwidth]{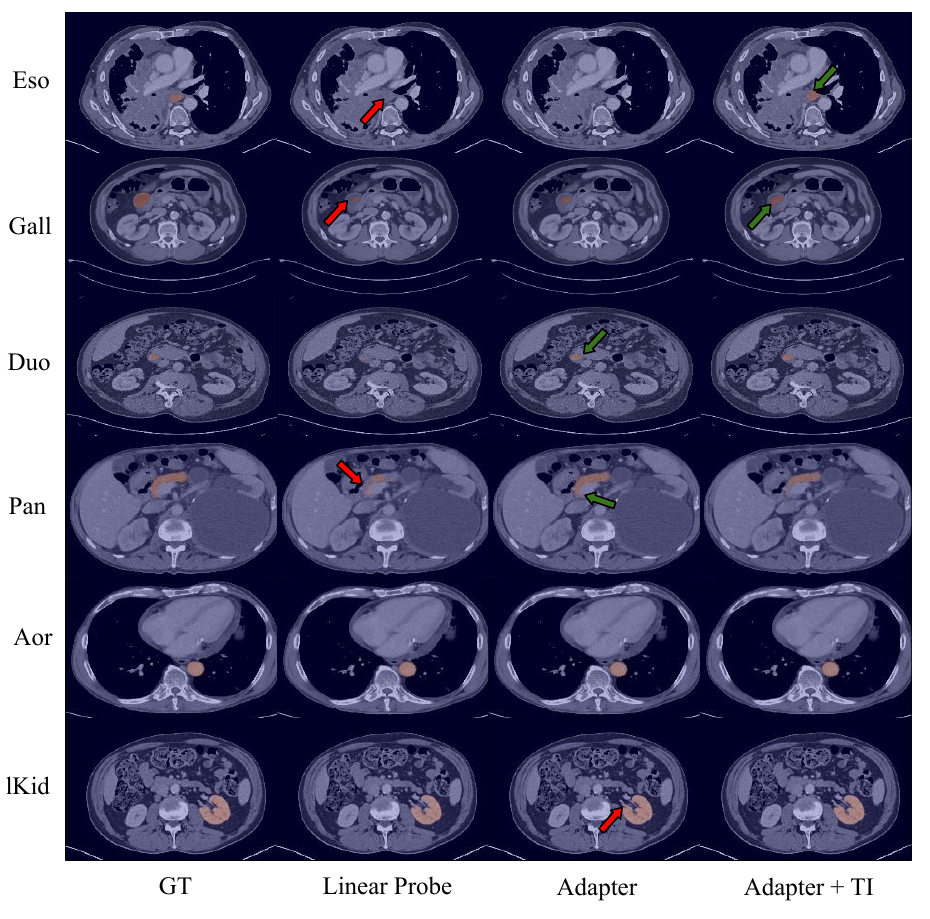}
\vspace{-2mm}
\caption{\textbf{Qualitative evaluation of the transductive inference}. Axial views of CT scans from TotalSegmentator dataset. The annotation/prediction masks of the target organ are in red. The effect of the black-box adaptation with K=5 shots in the presence/absence of transductive inference (TI) leveraging anatomic priors according to Section~\ref{subsec:method_ti} is illustrated.}
\label{fig:qualitative}
\end{figure}

\textit{\textbf{Qualitative evaluation}}. Figure~\ref{fig:qualitative} depicts a qualitative assessment of the performance of the proposed Adapter using the few-shot setting with k=5. The visualizations show the benefits of incorporating anatomical constraints regarding organ proportion during adaptation (\textit{first and second rows}). Also, the segmentation improvement of training a small Adapter module on top of the backbone for efficient transfer learning (\textit{second, third, and fourth rows}) is illustrated.

\begin{table}[h!]
\caption{\textbf{Leveraging anatomical priors during inference.} The transductive setting detailed in Section~\ref{subsec:method_ti} is applied using black-box Adapters in the low-data regime. Average results across nine organs. TI: transductive inference.}
\label{transductive_table}
\centering
\scriptsize
\begin{tabular}{lccc}
\toprule
\multicolumn{1}{c}{\multirow{1}{*}{Method}} & K=1     & K=5    & K=10 \\
\midrule
Linear probe            & 75.06 & 76.11 & 76.35 \\
Linear probe + TI       & \textbf{75.47}\improvement{0.41} & \textbf{76.67}\improvement{0.56} & \textbf{76.96}\improvement{0.61} \\
\hdashline\noalign{\vskip 0.5ex}
Spatial Adapter                 & 76.25 & 76.78 & 77.35 \\
Spatial Adapter + TI            & \textbf{77.11}\improvement{0.86} & \textbf{77.44}\improvement{0.66} & \textbf{77.99}\improvement{0.64} \\
\bottomrule
\end{tabular}
\end{table}

\subsubsection{Transferability to novel classes}
\label{subsec:novel}

A desirable quality of transfer learning is to leverage the learned universal representations in new concepts. In particular, we argue that such a property is of interest if the foundation model requires a small number of examples for adaptation. In the following, we evaluate the capabilities of our supervised pre-trained model in such a task by adapting it to the parcellation of three novel organs (including 16 anatomical structures) using K=10 support volumes. The results are presented in Table~\ref{results_our_novel}, and the main observations are given below.

\textit{\textbf{Comparing pre-training strategies}}. One can observe the performance improvement when fine-tuning the supervised pre-trained foundation model compared to its self-supervised counterpart from \cite{SwinUNETRweights} (\textit{see} Table~\ref{results_our_novel}, \textit{top}). Concretely, average gains of $+9.7\%$ are achieved using our developed foundation model. This suggests that pre-training for the underlying objective tasks, i.e., segmentation, might be more effective than pre-training on customized pretext tasks. This is especially the case when considering the pre-training data sources of self-supervised pre-training, e.g., \cite{SwinUNETRweights} weights are pre-trained on 5,050 unlabeled CT scans.

\textit{\textbf{Efficient adaptation methods}}. The direct application of PEFT and black-box adaptation strategies to novel tasks does not offer the same benefits as in known tasks. All these efficient alternatives perform below fine-tuning in this case. We argue that this phenomenon derives from freezing the decoder network, which is specialized in reconstructing the base categories masks. Indeed, fine-tuning only the decoder already provides similar results as full fine-tuning ($-0.4\%$). Thus, as expected, black-box Adapters are not a competitive solution in this setting. In addition, this limitation hinders a computationally efficient transfer since the most advanced volumetric segmentation network decoders contain a relatively large number of parameters, as we later elaborate in Section~\ref{comp_efficiency}. On the other hand, the results showcase that combining PEFT with decoder fine-tuning is an appealing alternative for data-efficient transfer learning to novel tasks performance-wise. For example, combining decoder fine-tuning with LoRA brings average performance gains of $+5.0\%$ compared to full fine-tuning.

\begin{table}[t!]
\setlength{\tabcolsep}{2.8pt}
\centering
\caption{\textbf{Transferability to new tasks on TotalSegmentator dataset}. Our foundation model is adapted for each organ region using K=10 support volumes for training. The metric presented is the Dice similarity coefficient averaged by organs. The best method results are highlighted in bold, and the second-best performance is underscored. FULL: Whole model training; PEFT: Parameter-Efficient Fine-Tuning; BB: black-box.} \label{results_our_novel}
\scriptsize
\begin{tabular}{llcccc}
\toprule
\multicolumn{1}{c}{Setting} & \multicolumn{1}{c}{Method}                     & Lung$^{*}$   & Heart$^{\dagger}$  & Gluteus$^{\ddagger}$  & \textbf{Avg.}  \\ \midrule
\multirow{2}{*}{FULL}           & Fine-tuning \citep{SwinUNETRweights}           & 19.59 & 53.14 & 55.37  & 42.70 \\ 
                                & Fine-tuning (\textit{Ours})                    & \textbf{31.01} & 60.79 & 65.35 & 52.38 \\  \cdashlinelr{1-6}
\multirow{9}{*}{PEFT}           & BitFit \citep{BenZaken2021BitFit}              & 14.79 & 48.90 & 39.43  & 34.28 \\ 
                                & LoRA \citep{hu2022lora}                        & 13.80 & 50.55 & 46.36  & 38.49 \\ 
                                & AdaptFormer \citep{adaptformer}                & 18.82 & 53.35 & 48.61  & 40.26 \\ 
                                & Affine-LN \citep{layernorm}                    & 16.92 & 58.38 & 46.07  & 40.46 \\
                                & Decoder fine-tuning                            & 25.98 & \underline{65.69} & 64.23  & 51.97 \\
                                &\hspace{2mm}+BitFit \citep{BenZaken2021BitFit}  & \underline{26.17} & 65.78 & 64.34 & 52.10 \\
                                &\hspace{2mm}+LoRA \citep{hu2022lora}            & 26.16 & \textbf{76.12} & \textbf{69.89} & \textbf{57.39} \\
                                &\hspace{2mm}+AdaptFormer \citep{adaptformer}    & 23.84 & 72.32 & \underline{69.79} & \underline{55.32} \\
                                &\hspace{2mm}+Affine-LN \citep{layernorm}        & 26.09 & 65.91 & 64.53 & 52.18 \\
\cdashlinelr{1-6}
\multirow{2}{*}{BB}             & Linear probe                                 &  9.35 &  9.19 &  7.52 &  8.68 \\ 
                                & Spatial Adapter                              & 10.08 & 14.66 & 12.75 & 12.50 \\ 
\bottomrule
\multicolumn{6}{l}{$^{*}$ Avg. of five: upper/lower lobe left, upper/lower lobe right, middle lobe.} \\
\multicolumn{6}{l}{$^{\dagger}$ Avg. of five: myocardium, atrium/ventricle left, atrium/verticle right.} \\
\multicolumn{6}{l}{$^{\ddagger}$ Avg. of six: maximus left/right, medius left/right, minimus left/right.} \\
\end{tabular}
\end{table}

\subsection{Transferrability on FLARE'22}

\begin{table*}[h!]
\centering
\caption{\textbf{Few-shot efficient adaptation results using the proposed foundation model (\textit{Ours}) on FLARE'22}. Adaptation is performed for all organs simultaneously, following a multi-class training. The metric presented is Dice. The best method results are highlighted in bold, and the second-best performance is underscored. FULL: Whole model training; PEFT: Parameter-Efficient Fine-Tuning; BB: black-box. The proposed black-box Adapter performance is shadowed.} \label{results_ours_flare}
\scriptsize
\begin{tabular}{cllcccccccccc}
\toprule
\multicolumn{2}{c}{Setting} & \multicolumn{1}{c}{Method}                                             & Spl   & lKid  & Gall  & Eso   & Liv   & Pan   & Sto   & Duo   & Aor   & \textbf{Avg.}  \\ \midrule
&                                                         & Zero-shot                           & 88.97 & 75.52 & 54.78 & 75.56 & 94.58 & 84.06 & 79.38 & 23.52 & 86.90 & 73.70 \\ 
\midrule 
\multirow{8}{*}{1-shot} & \multirow{2}{*}{FULL}           & Fine-tuning \citep{SwinUNETRweights}     & 12.62 &  8.94 & \multicolumn{1}{r}{4.47} &  \multicolumn{1}{r}{0.02} & 38.14 &  \multicolumn{1}{r}{3.68} &  \multicolumn{1}{r}{6.63} &  \multicolumn{1}{r}{2.69} &  \multicolumn{1}{r}{2.40} &  \multicolumn{1}{r}{8.84} \\
&                                                         & Fine-tuning (\textit{Ours})              & 84.98 & 73.06 & 54.85 & 71.64 & 94.54 & 82.15 & 74.69 & 42.75 & 91.01 & 74.41 \\ \cdashlinelr{2-13}
& \multirow{4}{*}{PEFT}                                   & BitFit \citep{BenZaken2021BitFit}        & 86.48 & 76.09 & 54.48 & 72.95 & 94.97 & 81.42 & 73.21 & 43.52 & 90.91 & 74.89 \\
&                                                         & LoRA \citep{hu2022lora}                  & 87.81 & 76.22 & 55.09 & 74.33 & 95.05 & 82.60 & 74.72 & 36.03 & 90.31 & 74.68 \\
&                                                         & AdaptFormer \citep{adaptformer}          & 83.69 & 76.77 & 54.49 & 72.13 & 94.24 & 80.01 & 73.85 & 43.86 & 90.79 & 74.43 \\
&                                                         & Affine-LN \citep{layernorm}              & 85.46 & 76.55 & 55.91 & 72.23 & 94.85 & 82.20 & 73.14 & 44.66 & 91.57 & 75.17 \\ \cdashlinelr{2-13}
& \multirow{2}{*}{BB}                                     & Linear probe                             & 88.42 & 75.00 & 56.08 & 74.71 & 95.02 & 82.82 & 78.14 & 42.01 & 88.88 & \underline{75.68} \\
&                                                         & \our Spatial Adapter                     & \our88.65 & \our75.31 & \our55.95 & \our74.53 & \our94.81 & \our83.55 & \our78.40  & \our43.99 & \our88.89 & \our \textbf{76.01} \\
\midrule 
\multirow{8}{*}{5-shot} & \multirow{2}{*}{FULL}           & Fine-tuning \citep{SwinUNETRweights}     & 23.20 & \multicolumn{1}{r}{8.72}  & 32.30 & 14.31 & 37.46 & 11.61 & 10.00 &  \multicolumn{1}{r}{6.99} & 42.90 & 20.84 \\
&                                                         & Fine-tuning (\textit{Ours})              & 86.89 & 75.66 & 53.67 & 73.58 & 95.45 & 82.46 & 79.18 & 54.48 & 92.06 & \textbf{77.05} \\ \cdashlinelr{2-13}
& \multirow{4}{*}{PEFT}                                   & BitFit \citep{BenZaken2021BitFit}        & 88.24 & 77.16 & 55.23 & 74.93 & 95.71 & 83.20 & 77.58 & 45.87 & 91.05 & 76.55 \\
&                                                         & LoRA \citep{hu2022lora}                  & 86.56 & 76.15 & 54.24 & 74.46 & 94.93 & 83.01 & 75.17 & 43.66 & 92.01 & 75.58 \\
&                                                         & AdaptFormer \citep{adaptformer}          & 81.50 & 75.36 & 55.86 & 74.53 & 94.66 & 82.87 & 76.15 & 46.89 & 91.43 & 75.47 \\
&                                                         & Affine-LN \citep{layernorm}              & 85.73 & 76.00 & 53.71 & 75.2  & 94.62 & 82.04 & 74.69 & 47.24 & 91.38 & 75.62 \\ \cdashlinelr{2-13}
& \multirow{2}{*}{BB}                                     & Linear probe                             & 88.80 & 75.87 & 56.37 & 75.52 & 95.31 & 83.64 & 80.04 & 45.11 & 89.38 & 76.67 \\
&                                                         & \our Spatial Adapter                     & \our88.95 & \our74.23 & \our56.33 & \our74.34 & \our95.32 & \our83.94 & \our79.15 & \our48.59 & \our89.39 & \our \underline{76.69} \\
\midrule
\multirow{8}{*}{10-shot} & \multirow{2}{*}{FULL}          & Fine-tuning \citep{SwinUNETRweights}     & 47.39 & 30.06 & 40.09 & 15.53 & 60.59 & 34.99 & 21.22 & 25.16 & 67.21 & 38.03 \\
&                                                         & Fine-tuning (\textit{Ours})              & 80.56 & 77.24 & 54.27 & 76.11 & 92.91 & 82.50 & 78.73 & 56.39 & 92.01 & 76.75 \\ \cdashlinelr{2-13}
& \multirow{4}{*}{PEFT}                                   & BitFit \citep{BenZaken2021BitFit}        & 86.91 & 76.65 & 55.87 & 75.36 & 95.08 & 82.83 & 77.30 & 49.24 & 91.43 & 76.74 \\
&                                                         & LoRA \citep{hu2022lora}                  & 84.00 & 77.05 & 54.72 & 74.02 & 94.64 & 82.66 & 76.31 & 46.11 & 91.59 & 75.68 \\
&                                                         & AdaptFormer \citep{adaptformer}          & 87.08 & 76.96 & 55.69 & 74.78 & 93.69 & 82.45 & 77.07 & 46.29 & 91.38 & 76.15 \\
&                                                         & Affine-LN \citep{layernorm}              & 87.68 & 76.99 & 55.91 & 75.61 & 95.04 & 83.57 & 79.27 & 51.88 & 91.69 & \textbf{77.52} \\ \cdashlinelr{2-13}
& \multirow{2}{*}{BB}                                     & Linear probe                             & 88.80 & 75.50 & 56.21 & 74.75 & 95.38 & 83.92 & 80.02 & 45.21 & 89.42 & 76.58 \\
&                                                         & \our Spatial Adapter                     & \our89.58 & \our75.80 & \our56.31 & \our74.43 & \our95.35 & \our83.89 & \our79.59 & \our49.27 & \our89.84 & \our \underline{77.12} \\
\midrule 
\multirow{8}{*}{30-shot} & \multirow{2}{*}{FULL}          & Fine-tuning \citep{SwinUNETRweights}     & 67.55 & 63.31 & 42.88 & 28.60 & 83.81 & 62.11 & 58.63 & 45.61 & 82.03 & 59.39 \\
&                                                         & Fine-tuning (\textit{Ours})              & 87.95 & 75.82 & 55.37 & 77.65 & 95.68 & 82.11 & 79.97 & 60.71 & 92.29 & \textbf{78.62} \\ \cdashlinelr{2-13}
& \multirow{4}{*}{PEFT}                                   & BitFit \citep{BenZaken2021BitFit}        & 87.93 & 78.06 & 56.54 & 75.17 & 95.32 & 83.58 & 79.88 & 50.84 & 92.38 & 77.74 \\
&                                                         & LoRA \citep{hu2022lora}                  & 87.89 & 77.79 & 55.67 & 75.41 & 94.88 & 84.16 & 78.16 & 49.95 & 92.08 & 77.33 \\
&                                                         & AdaptFormer \citep{adaptformer}          & 77.02 & 72.25 & 59.87 & 75.54 & 84.47 & 83.17 & 75.15 & 50.78 & 92.60 & 74.54 \\
&                                                         & Affine-LN \citep{layernorm}              & 88.31 & 78.00 & 54.90 & 72.86 & 95.60 & 84.22 & 79.02 & 56.61 & 92.11 & \underline{77.96} \\ \cdashlinelr{2-13}
& \multirow{2}{*}{BB}                                     & Linear probe                             & 88.97 & 75.40 & 56.55 & 74.26 & 95.44 & 83.86 & 79.42 & 46.68 & 90.02 & 76.73 \\
&                                                         & \our Spatial Adapter                     & \our89.55 & \our75.54 & \our56.48 & \our75.34 & \our95.50 & \our83.97 & \our80.34 & \our51.06 & \our90.66 & \our77.60 \\
\bottomrule
\end{tabular}
\end{table*}

\begin{figure*}[ht!]
\centering
\includegraphics[width=0.9\textwidth]{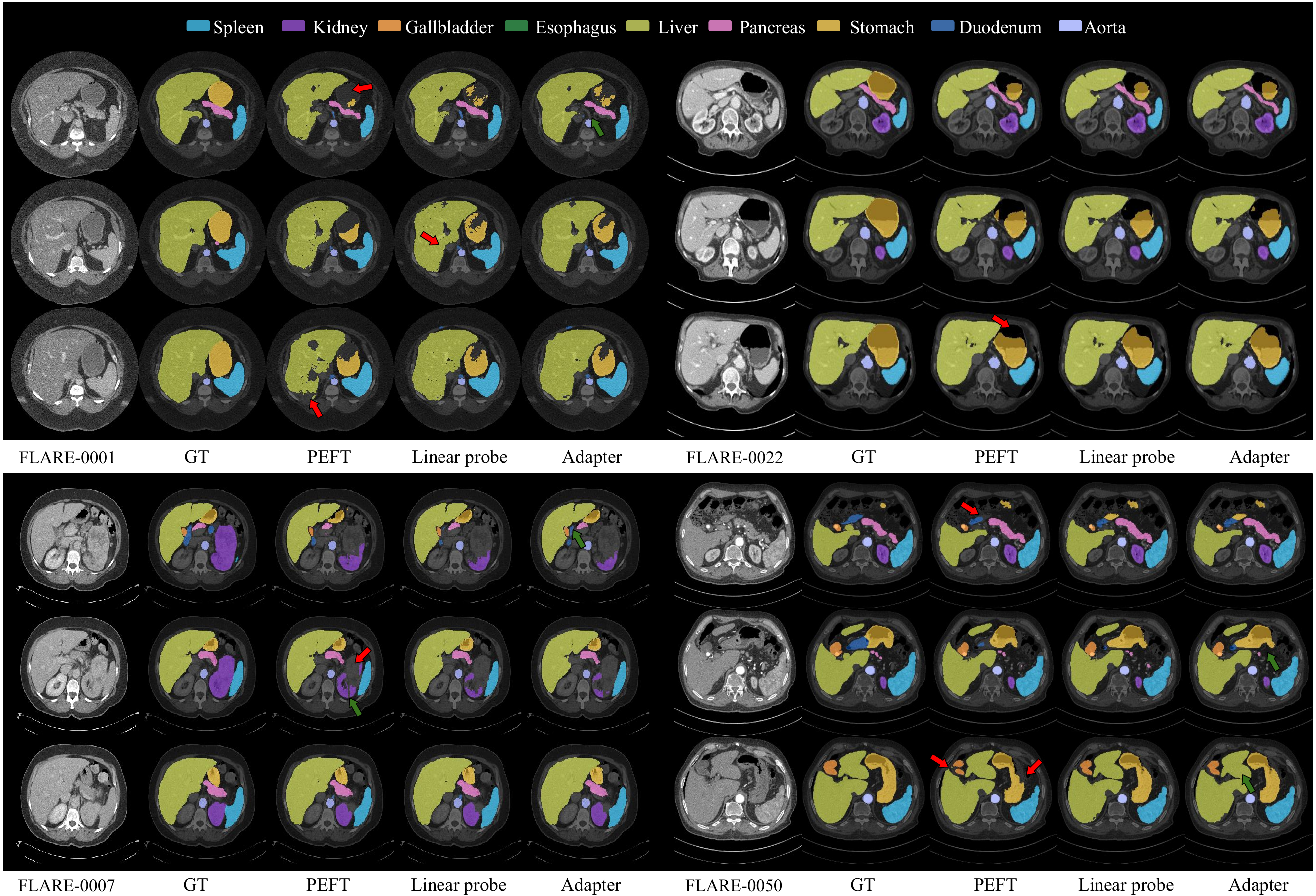}
\caption{\textbf{Few-shot adaptation on FLARE'22}. The capabilities of PEFT and black-box Adapters for transferring the proposed supervised pre-trained foundation model are illustrated. Results using K=10 annotated support volumes for adaptation to the segmentation of nine base organs with domain shifts.}
\label{fig:qualitative_flare}
\end{figure*}

The results transferring the pre-trained foundation model on the FLARE'22 dataset are introduced in Table~\ref{results_ours_flare}. Also, a graphical representation of such results is introduced in Figure~\ref{fig:intro}(a), and qualitative examples are displayed in Figure~\ref{fig:qualitative_flare}. The main findings from these experiments are detailed below.

\textit{\textbf{Zero-shot}}. The results obtained when directly evaluating the pre-trained model on this external dataset are similar to the ones obtained in TotalSegmentator, reaching an average DSC of 73.70, which shows the robustness of large-scale supervised pre-training to generalize across domains. Worthy of note, the results are above and beyond fine-tuning from self-supervised pre-trained weights in \cite{SwinUNETRweights} using K=30 shots.

\textit{\textbf{Large data regime}}. Results with K=30 show that fine-tuning is a more appealing alternative performance-wise than PEFT in this case. Regarding the black-box Adapters, the proposed Spatial Adapter tailored for dense predictions offers significant gains compared to linear probing ($+0.9\%$).

\textit{\textbf{Low data regime}}. FLARE'22 transferability experiments using K=$\{1, 5, 10\}$ provide similar impressions if compared to TotalSegmentator results. The proposed Spatial black-box Adapter again brings consistent gains over linear probing, and is competitive compared to PEFT methods. This is especially the case for K=1, where full fine-tuning and PEFT consistently drop their performance. It should be noted that FLARE'22 poses a larger domain shift than TotalSegmentator, mainly caused by the diseases present in the former's scans. As illustrated in Figure~\ref{fig:qualitative_flare} (e.g., \textit{bottom-left scan}), there exist anatomical structures with anomalies. Hence, the pre-trained model might require more severe modifications to accommodate the annotation bias of such structures.

\subsection{Computational Efficiency}
\label{comp_efficiency}

Concrete figures of merit that support the computational efficiency of the proposed adaptation strategies are introduced in Table~\ref{computation_eficiency}, using the proposed foundation model that builds upon the Swin-UNETR architecture. Additionally, the performance \textit{vs.} efficiency trade-off for each strategy can be further visualized in Figure~\ref{fig:intro}(b). The efficiency of the adaptation strategies is evaluated with two key performance indicators: the number of parameters required for tuning and the overall training time.

\textit{\textbf{Effect of pre-training strategy}}. Even though two strategies might introduce the same number of learnable parameters, training times vary depending on the domain, task shift, and early stopping criteria. For example, one can notice shorter training times when full fine-tuning using supervised pre-training compared to its self-supervised counterpart, despite both updating the same number of weights.

\textit{\textbf{Parameter-Efficient Fine-Tuning}}. PEFT methods stand out by their reduced training times (around five minutes for TotalSegmentator experiments) and reduced number of tuning parameters compared to full fine-tuning. It is noticeable that decoder fine-tuning incorporates a notorious number of tuning parameters (19.6M) and longer training times, so this strategy is arguably closer to full fine-tuning approaches. Therefore, we argue that keeping this component frozen is paramount to ensure a computationally efficient transfer.

\textit{\textbf{Black-box Adapters}}. Despite introducing more trainable parameters than PEFT strategies, such as LoRA or LayerNorm tuning, the proposed Spatial Adapter requires similar processing times while yielding significant performance improvements over these approaches (\textit{see} Figure~\ref{fig:intro}(b)), and operate in black-box scenarios. Finally, a simple linear probe constitutes the most efficient adaptation strategy.

\textit{\textbf{Differences between datasets}}. Longer training times are observed in FLARE'22, which are explained by slower convergence due to the multi-class nature of the task.

\begin{table}[h!]
\setlength{\tabcolsep}{2pt}
\centering
\caption{\textbf{Computational efficiency analysis}. Trainable parameters ($\#$Param.) and overall training time (T) of different PEFT methods and black-box Adapters, using our foundation model on base tasks. Training times are obtained using K=5 support volumes. Results are averaged across nine organs for TotalSegmentator, and reported in the multi-class scenario for FLARE'22. The adaptation is done on a single A6000 GPU, following the implementation details described in \ref{subsec:adapting_fm}. PEFT: Parameter-Efficient Fine-Tuning; BB: black-box.}
\label{computation_eficiency}
\scriptsize{
\begin{tabular}{lcrrrr}
\toprule
\multicolumn{1}{c}{Method} & \multicolumn{1}{c}{Category} & \multicolumn{2}{c}{TotalSegmentator} & \multicolumn{2}{c}{FLARE'22} \\ \midrule
 & & \multicolumn{1}{c}{$\#$Param.} & \multicolumn{1}{c}{T(min)} & \multicolumn{1}{c}{$\#$Param.} & \multicolumn{1}{c}{T(min)} \\ \midrule
Fine-tuning \citep{SwinUNETRweights}                                     & \multirow{2}{*}{FULL}         & 62.1M   &  15  & 62.1M   &  50 \\
Fine-tuning (\textit{Ours})                                 &                               & 62.1M   &   8  & 62.1M   &  35 \\ \noalign{\vskip 0.5ex}\cdashlinelr{1-6}\noalign{\vskip 0.5ex}
Decoder                                     & \multirow{5}{*}{PEFT}         & 19.6M   &   7  & 19.6M   &  32 \\
Bitfit \citep{BenZaken2021BitFit}           &                               & 210.7K  &   5  & 211.1K  &  29 \\
LoRA \citep{hu2022lora}                     &                               & 68.1K   &   6  & 69.4K   &  25 \\
AdaptFormer \citep{adaptformer}             &                               & 47.6K   &   7  & 48.1K   &  24 \\
Affine-LN \citep{layernorm}                 &                               & 17.3K   &   5  & 17.7K   &  25 \\ \noalign{\vskip 0.5ex}\cdashlinelr{1-6}\noalign{\vskip 0.5ex}
Linear probe                                & \multirow{2}{*}{BB}           & 49      &   4  & 490     &  7  \\ 
Spatial Adapter                             &                               & 124.4K  &   5  & 124.9K  &  11 \\ \bottomrule 
\end{tabular}
}
\end{table}

\subsection{Adapting open-access available pre-trained models}

The underlying philosophy of the pre-training and adaptation paradigm is to enable the efficient transfer of large-scale open-access models. To assess the agnostic nature of the proposed setting for any large-scale pre-trained network, this section explores its impact on publicly available supervised pre-trained models. More concretely, three versions of the same Swin-UNETR used in this work and one purely convolutional U-Net model are considered. First, a dataset-specific model with weights released by the authors in \cite{SwinUNETRweights} is used. This model was trained on 24 volumes and 13 abdominal organs, uniquely from the BTCV \citep{BTCV} dataset. Second, a recently released large-scale foundation model for volumetric organ and lesion segmentation, the so-called CLIP-Driven model \citep{UniversalModel}, is employed. This model was trained on 2,100 CT volumes, 32 target tasks, and ten datasets. The datasets used in \cite{UniversalModel} coincide with the ones in Table~\ref{datasets}, and include additional data from WORD \citep{luo2022word}. Last, the concurrent work of \cite{suprem} is considered, which provides a set of supervised pre-trained models, SuPreM, using the same scans as in the CLIP-Driven model but refining labels through active learning. Both the Swin-UNETR and 3D-Unet (convolutional) models are selected. The experiments are carried out on TotalSegmentator base tasks, which have been previously employed by \cite{UniversalModel} and \cite{suprem} for evaluating the transferability of their foundation models. In the following, the main findings learned from the adaptation of these models are elaborated.

\textit{\textbf{Dataset-specific pre-training}}. Results in Table~\ref{results_btcv} point out the necessity of developing open-access models trained on various datasets and tasks. The dataset-specialized model from \cite{SwinUNETRweights} shows poor transferability capabilities in the low-data regime. Concretely, using K=5, linear probing degrades the performance by nearly $27\%$ compared to the large-scale foundation model trained in this work.

\textit{\textbf{CLIP-Driven model}}. First, results in Table~\ref{results_clipdriven} show similar performance between the CLIP-Driven model in \cite{UniversalModel} and the foundation model presented in this work. This observation questions the utility of text information in the context of organ segmentation and a narrow set of concepts. Second, it is worth noting that black-box adaptation methods show again better transferability than PEFT approaches in the low data regime (e.g., when K=5). Last, using the proposed Spatial Adapter brings consistent average improvements over linear probing, ranging $[+0.2, +3.6]\%$.

\textit{\textbf{SuPreM models}}. Results in Table~\ref{results_suprem_totalseg} using \cite{suprem} models show the best few-shot transferability of all considered networks. Interestingly, black-box Adapters outperform PEFT methods. This is especially the case for the 3D-Unet model (\textit{see} Table~\ref{results_suprem_totalseg}(a)), for which PEFT methods struggle to perform properly. In contrast, PEFT methods in the Transformer architecture (\textit{see} Table~\ref{results_suprem_totalseg}(b)) present more robust results. Comparing PEFT and black-box Adapters, the results suggest that \textit{better pre-trained foundation models also lead to relatively stronger black-box adaptation}. Finally, the performance gains derived from Spatial Adapters are worth noting. The proposed black-box method improves linear probing $+6.3\%$ for K=5 and $+8.0\%$ for K=10 in the case of 3D-Unet, and average improvements up to $+1.4\%$ are observed in Swin-UNETR.

\begin{table*}[h!]
\centering
\caption{\textbf{Few-shot efficient adaptation results using a dataset-specific model from \cite{SwinUNETRweights} on TotalSegmentator}. In this setting, the pre-trained model used for adaptation was trained on a single dataset (BTCV in Table~\ref{datasets}) containing 24 labeled volumes. Adaptation is performed for each organ individually in a binary segmentation task. The metric presented is Dice. The best method results are highlighted in bold, and the second-best performance is underscored. PEFT: Parameter-Efficient Fine-Tuning; BB: black-box. The proposed black-box Adapter performance is shadowed.}
\label{results_btcv}
\scriptsize{
\begin{tabular}{cllccccccccc}
\toprule
\multicolumn{2}{c}{Setting} & \multicolumn{1}{c}{Method}                                                         & Spl    & lKid   & Gall  & Eso   & Liv   & Pan   & Sto    & Aor  & \textbf{Avg.}$^{*}$  \\ \midrule
\multirow{6}{*}{5-shot} & \multirow{4}{*}{PEFT}                 & BitFit \citep{BenZaken2021BitFit}              & 67.58  & 33.73  & 47.65 & 43.48 & 79.41 & 33.65 & 41.75 & 70.16 & \underline{52.18} \\ 
&                                                               & LoRA \citep{hu2022lora}                        & 74.05  & 28.86  & 30.81 & 42.20 & 62.89 & 40.74 & 41.74 & 68.23 & 48.69 \\ 
&                                                               & AdaptFormer \citep{adaptformer}                & 71.58  & 34.79  & 54.52 & 43.30 & 65.39 & 41.11 & 44.58 & 70.81 & \textbf{53.26} \\
&                                                               & Affine-LN \citep{layernorm}              & 71.13  & 37.85  & 46.08 & 46.20 & 69.19 & 38.44 & 37.44 & 70.78 & 52.14 \\ \cdashlinelr{2-12}
& \multirow{2}{*}{BB}                                           & Linear probe                                   & 56.60  & 44.95  & 39.80 & 37.98 & 58.94 & 43.66 & 44.15 & 65.83 & 48.99 \\
&                                                               & \our Spatial Adapter                           & \our57.45  & \our40.36  & \our35.80 & \our42.56 & \our59.15 & \our47.38 & \our44.64 & \our65.45 & \our49.10 \\
\midrule
\multirow{6}{*}{10-shot} & \multirow{4}{*}{PEFT}                & BitFit \citep{BenZaken2021BitFit}              & 82.83  & 25.99  & 55.13 & 47.21 & 80.05 & 44.44 & 43.88 & 85.05 & 58.07 \\
&                                                               & LoRA \citep{hu2022lora}                        & 71.52  & 45.74  & 38.85 & 46.86 & 78.25 & 53.55 & 47.03 & 85.37 & \underline{58.40} \\
&                                                               & AdaptFormer \citep{adaptformer}                & 79.75  & 42.98  & 55.34 & 50.28 & 79.30 & 48.25 & 51.40 & 85.98 & \textbf{61.66} \\
&                                                               & Affine-LN \citep{layernorm}              & 75.54  & 40.08  & 52.26 & 40.35 & 80.31 & 47.78 & 48.65 & 81.51 & 58.31 \\   \cdashlinelr{2-12}
& \multirow{2}{*}{BB}                                           & Linear probe                                   & 61.26  & 42.46  & 39.63 & 37.69 & 46.71 & 36.61 & 44.05 & 65.45 & 46.73 \\
&                                                               & \our Spatial Adapter                           & \our60.59  & \our39.52  & \our42.83 & \our43.49 & \our58.78 & \our36.62 & \our45.88 & \our66.99 & \our49.34 \\
\bottomrule
\multicolumn{12}{l}{$^{*}$ Since the duodenum was not used for pre-training, this organ was not included to ensure fair comparisons on base organs.} \\
\end{tabular}
}
\end{table*}

\begin{table*}[h!]
\centering
\caption{\textbf{Few-shot efficient adaptation results using CLIP-Driven model \citep{UniversalModel} on TotalSegmentator}. In this setting, the pre-trained model used for adaptation was trained on ten diverse datasets containing 2,100 labeled volumes, with training data largely overlapping Table~\ref{datasets} datasets. Adaptation is performed for each organ individually in a binary segmentation task. The metric presented is Dice. The best method results are highlighted in bold, and the second-best performance is underscored. PEFT: Parameter-Efficient Fine-Tuning; BB: black-box. The proposed black-box Adapter performance is shadowed.}
\label{results_clipdriven}
\scriptsize{
\begin{tabular}{cllcccccccccc}
\toprule
\multicolumn{2}{c}{Setting} & \multicolumn{1}{c}{Method}                                                         & Spl    & lKid  & Gall  & Eso   & Liv   & Pan   & Sto   & Duo   & Aor   & \textbf{Avg.}    \\ \midrule
\multirow{6}{*}{5-shot} & \multirow{4}{*}{PEFT}                 & BitFit \citep{BenZaken2021BitFit}              & 96.02 & 82.48 & 66.62 & 45.36 & 91.94 & 75.85 & 61.36 & 60.32 & 88.76 & 74.30  \\
&                                                               & LoRA \citep{hu2022lora}                        & 91.30 & 57.26 & 52.18 & 48.40 & 88.67 & 77.11 & 51.46 & 67.04 & 83.69 & 68.57 \\
&                                                               & AdaptFormer \citep{adaptformer}                & 93.35 & 75.20 & 56.44 & 48.16 & 92.30 & 73.23 & 55.27 & 66.36 & 89.61 & 72.21 \\ 
&                                                               & Affine-LN \citep{layernorm}                    & 95.89 & 84.61 & 70.33 & 44.80 & 91.87 & 76.66 & 58.73 & 54.56 & 89.53 & 74.11 \\ \cdashlinelr{2-13}
& \multirow{2}{*}{BB}                                           & Linear probe                                   & 94.37 & 86.52 & 60.95 & 53.49 & 90.53 & 75.50 & 60.20 & 74.07 & 80.75 & \underline{75.15}  \\
&                                                               & \our Spatial Adapter                           & \our96.40 & \our86.79 & \our66.84 & \our51.20 & \our92.53 & \our73.13 & \our62.84 & \our73.76 & \our74.69 & \our \textbf{75.35}  \\
\midrule
\multirow{6}{*}{10-shot} & \multirow{4}{*}{PEFT}                & BitFit \citep{BenZaken2021BitFit}              & 90.49 & 87.32 & 72.06 & 51.98 & 92.82 & 78.49 & 59.81 & 68.51 & 90.13 & \underline{76.85} \\
&                                                               & LoRA \citep{hu2022lora}                        & 72.48 & 71.11 & 56.51 & 53.29 & 90.18 & 82.62 & 56.94 & 73.15 & 86.40  & 71.41 \\
&                                                               & AdaptFormer \citep{adaptformer}                & 89.54 & 76.64 & 66.90  & 52.42 & 92.88 & 72.87 & 71.67 & 68.60  & 91.15 & 75.85 \\
&                                                               & Affine-LN \citep{layernorm}                    & 95.24 & 83.92 & 75.89 & 51.57 & 92.33 & 75.62 & 61.24 & 73.77 & 90.43 & \textbf{77.78} \\  \cdashlinelr{2-13}
& \multirow{2}{*}{BB}                                           & Linear probe                                   & 96.19 & 88.19 & 57.83 & 53.25 & 87.20  & 66.69 & 58.09 & 65.29 & 81.22 & 72.66 \\
&                                                               & \our Spatial Adapter                           & \our96.58 & \our86.88 & \our67.08 & \our54.07 & \our92.36 & \our72.26 & \our62.95 & \our74.61 & \our79.28 & \our76.23 \\
\bottomrule
\end{tabular}
}
\end{table*}

\begin{table*}[h!]
\centering
\caption{\textbf{Few-shot efficient adaptation results transferring SuPreM models \citep{suprem} on TotalSegmentator}. Such a family of networks is pre-trained on a diverse dataset assembly containing 2,100 labeled volumes, with labels refined through an active learning pipeline. Adaptation is performed for each organ individually in a binary segmentation task. The metric presented is Dice. The best method results are highlighted in bold, and the second-best performance is underscored. PEFT: Parameter-Efficient Fine-Tuning; BB: black-box. The proposed black-box Adapter performance is shadowed.} 
\label{results_suprem_totalseg}

\centering
\scriptsize{
\begin{tabular}{cllcccccccccc}
\toprule
\multicolumn{2}{c}{Setting} & \multicolumn{1}{c}{Method}                                                    & Spl   & lKid  & Gall  & Eso   & Liv   & Pan   & Sto   & Duo   & Aor   & \textbf{Avg.}  \\ \midrule
\multirow{6}{*}{5-shot} & \multirow{3}{*}{PEFT}                 & CNN-Adapter \citep{Rebuffi2018}           & 47.69 & 39.58 & 40.52 & 53.05 & 55.08 & 43.17 & 28.47 & 35.73 & 84.62 & 47.55 \\
&                                                               & Bias \citep{tinytl}                       & 71.16 & 69.54 & 70.16 & 55.86 & 71.03 & 79.60 & 51.25 & 69.04 & 88.92 & 69.62 \\
&                                                               & Affine-BN \citep{frankle2021training}     & 69.22 & 72.33 & 65.66 & 52.68 & 67.61 & 75.50 & 45.08 & 66.52 & 86.94 & 66.84 \\ \cdashlinelr{2-13}
& \multirow{2}{*}{BB}                                           & Linear probe                              & 93.91 & 75.59 & 75.94 & 50.50 & 80.29 & 68.19 & 57.18 & 77.18 & 88.48 & \underline{74.14} \\
&                                                               & \our Spatial Adapter                      & \our91.78 & \our77.71 & \our80.89 & \our52.30 & \our90.00 & \our78.83 & \our83.27 & \our80.37 & \our89.08 & \our \textbf{80.47} \\
\midrule
\multirow{6}{*}{10-shot} & \multirow{3}{*}{PEFT}                & CNN-Adapter \citep{Rebuffi2018}           & 57.32 & 61.79 & 42.96 & 55.61 & 52.21 & 52.77 & 39.96 & 34.97 & 89.26 & 54.09 \\
&                                                               & Bias \citep{tinytl}                       & 72.79 & 76.14 & 83.37 & 59.65 & 73.97 & 79.68 & 60.65 & 73.46 & 92.80 & \underline{74.72} \\
&                                                               & Affine-BN \citep{frankle2021training}     & 72.15 & 74.06 & 77.15 & 58.65 & 72.31 & 77.08 & 61.74 & 63.94 & 92.43 & 72.17 \\ \cdashlinelr{2-13}
& \multirow{2}{*}{BB}                                           & Linear probe                              & 91.22 & 75.63 & 77.48 & 50.02 & 80.87 & 69.17 & 56.28 & 77.63 & 85.29 & 73.73 \\
&                                                               & \our Spatial Adapter                      & \our95.40 & \our83.76 & \our81.29 & \our52.49 & \our90.75 & \our78.57 & \our81.97 & \our81.09 & \our90.33 & \our \textbf{81.74} \\
\bottomrule
\end{tabular}
}
\captionsetup{labelformat=empty}
\caption{\textbf{(a) 3D-UNet}}
\addtocounter{table}{-1}
\scriptsize{
\begin{tabular}{cllcccccccccc}
\toprule
\multicolumn{2}{c}{Setting} & \multicolumn{1}{c}{Method}                                                   & Spl   & lKid  & Gall  & Eso   & Liv   & Pan   & Sto   & Duo   & Aor   & \textbf{Avg.}  \\ \midrule
\multirow{6}{*}{5-shot} & \multirow{4}{*}{PEFT}                 & BitFit \citep{BenZaken2021BitFit}        & 88.76 & 85.91 & 79.42 & 50.22 & 92.17 & 73.64 & 62.81 & 69.30 & 90.82 & 77.01 \\
&                                                               & LoRA \citep{hu2022lora}                  & 61.31 & 46.52 & 52.50 & 46.43 & 80.50 & 66.86 & 38.66 & 54.15 & 73.33 & 57.81 \\
&                                                               & AdaptFormer \citep{adaptformer}          & 87.57 & 86.05 & 60.17 & 51.79 & 90.11 & 76.73 & 68.29 & 74.49 & 93.12 & 76.48 \\
&                                                               & Affine-LN \citep{layernorm}              & 88.14 & 83.81 & 76.10 & 50.04 & 91.89 & 75.46 & 64.41 & 71.91 & 90.91 & 76.96 \\ \cdashlinelr{2-13}
& \multirow{2}{*}{BB}                                           & Linear probe                             & 94.62 & 91.86 & 82.98 & 49.29 & 93.54 & 78.86 & 72.43 & 77.30 & 88.77 & \underline{81.07} \\
&                                                               & \our Spatial Adapter                     & \our95.34 & \our88.13 & \our85.08 & \our55.56 & \our94.27 & \our78.84 & \our75.33 & \our78.17 & \our87.40 & \our \textbf{82.01} \\
\midrule
\multirow{6}{*}{10-shot} & \multirow{4}{*}{PEFT}                & BitFit \citep{BenZaken2021BitFit}        & 95.16 & 86.54 & 84.86 & 56.93 & 93.58 & 72.03 & 69.26 & 75.47 & 90.44 & \underline{80.47} \\
&                                                               & LoRA \citep{hu2022lora}                  & 63.97 & 54.53 & 59.25 & 55.33 & 84.03 & 77.72 & 58.72 & 73.89 & 80.59 & 67.56 \\
&                                                               & AdaptFormer \citep{adaptformer}          & 91.36 & 84.03 & 77.78 & 54.10 & 93.14 & 76.05 & 70.08 & 77.58 & 93.25 & 79.71 \\
&                                                               & Affine-LN \citep{layernorm}              & 87.21 & 87.36 & 80.84 & 55.80 & 93.65 & 76.98 & 66.78 & 75.66 & 92.50 & 79.64 \\ \cdashlinelr{2-13}
& \multirow{2}{*}{BB}                                           & Linear probe                             & 95.26 & 91.63 & 82.15 & 52.69 & 93.37 & 69.93 & 71.70 & 77.20 & 88.70 & 80.29 \\
&                                                               & \our Spatial Adapter                     & \our95.83 & \our89.44 & \our81.61 & \our56.24 & \our94.40 & \our77.69 & \our76.03 & \our79.54 & \our84.66 & \our \textbf{81.72} \\
\bottomrule
\end{tabular}
}
\captionsetup{labelformat=empty}
\caption{\textbf{(b) Swin-UNETR}}
\addtocounter{table}{-1}
\end{table*}

\subsection{Ablation experiments}
\label{ssec:ablation}

This section includes ablation experiments that empirically motivate the design choices and hyper-parameters employed.

\textit{\textbf{Decoder fine-tuning}}. For transferability experiments to base organs, we decided to re-use the frozen decoder (\textit{see} Section~\ref{sec:peft_implementation}). This methodological choice allows both parameter-efficient and black-box adaptation. Nevertheless, transfer learning experiments to novel tasks (\textit{see} Section~\ref{subsec:novel}) have showcased a need to update this component. The ablation study in Table~\ref{decoder_table} suggests that decoder fine-tuning might be an appealing alternative to full fine-tuning also for base organs when K$\geq$10, but not in the lower data regimes. However, its combination with PEFT does not bring consistent performance gains for base organs if compared to black-box Adapters in Table~\ref{results_ours}.

\begin{table}[t!]
\setlength{\tabcolsep}{4pt}
\caption{\textbf{Performance of decoder fine-tuning.} The effect of combining PEFT on the Swin-UNETR encoder and fine-tuning the convolutional decoder for transfer learning is shown. Relative improvements are provided in parentheses. Results averaged across the nine base tasks on TotalSegmentator.}
\vspace{-2mm}
\label{decoder_table}
\centering
\scriptsize
\begin{tabular}{lccc}
\toprule
\multicolumn{1}{c}{\multirow{1}{*}{Method}} & K=1     & K=5    & K=10 \\
\midrule
\multicolumn{4}{l}{Supervised pre-training (\textit{Ours}).} \\  \hdashline\noalign{\vskip 0.5ex}
Full fine-tuning                                      & \textbf{75.90} & \textbf{77.15} & 76.93 \\
Decoder fine-tuning                                   & 71.51 & 76.31 & 78.51 \\
\hspace{2mm}+BitFit \citep{BenZaken2021BitFit}   & 71.36\worsening{0.15}   & 75.91\worsening{0.40} & 78.52\improvement{0.01} \\
\hspace{2mm}+LoRA \citep{hu2022lora}             & 67.67\worsening{3.84}   & 75.09\worsening{1.22} & 75.97\worsening{2.54} \\
\hspace{2mm}+AdaptFormer \citep{adaptformer}     & 71.31\worsening{0.20}   & 75.55\worsening{0.76} & 78.24\worsening{0.27} \\
\hspace{2mm}+Affine-LN \citep{layernorm}         & 71.73\improvement{0.22} & 75.85\worsening{0.46} & \textbf{78.81}\improvement{0.30} \\
\bottomrule
\end{tabular}
\end{table}

\begin{figure}[t!]
\setlength{\tabcolsep}{2pt}
    \centering
        
         \begin{tabular}{cc}
             \includegraphics[width=.48\linewidth]{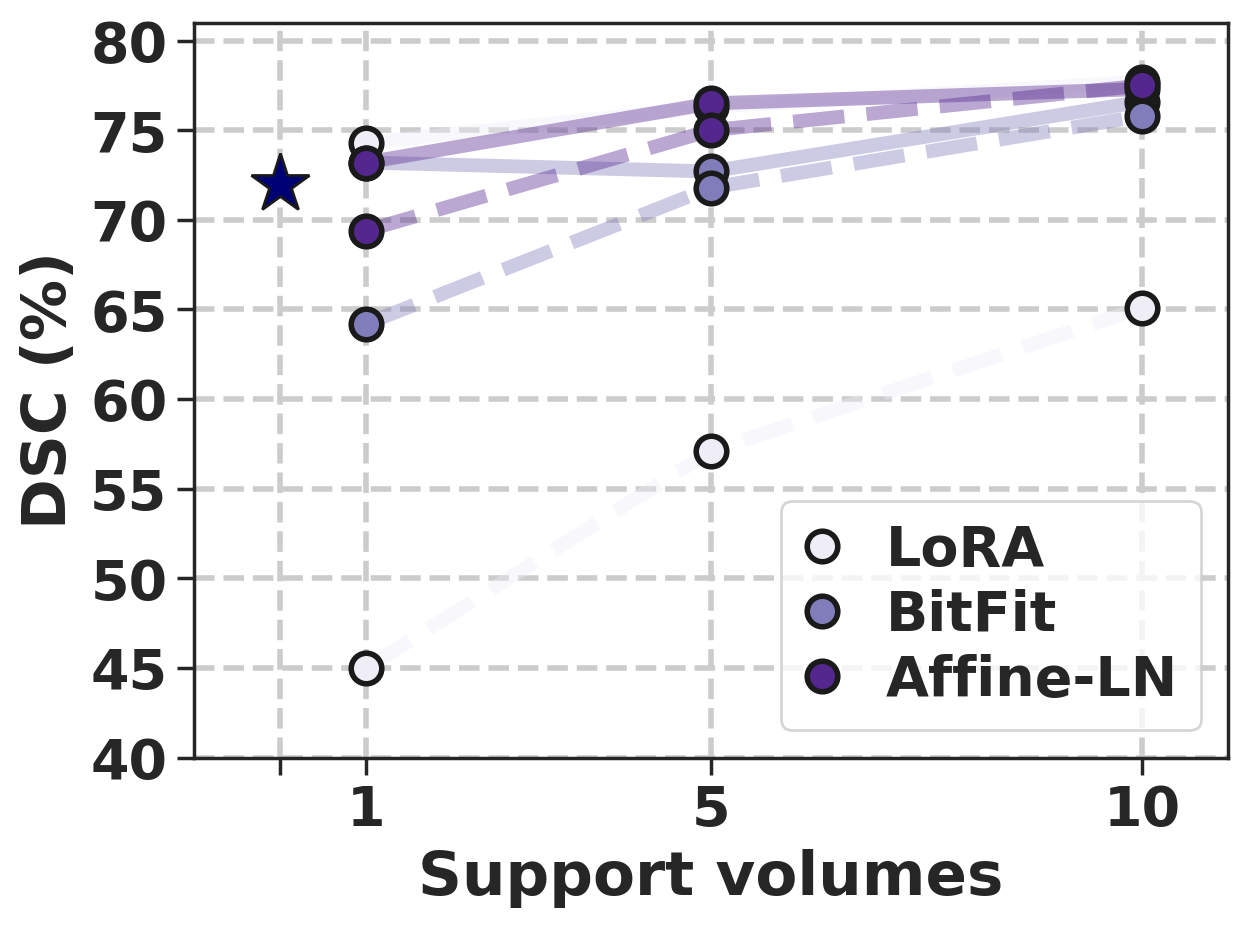} & 
            \includegraphics[width=.48\linewidth]{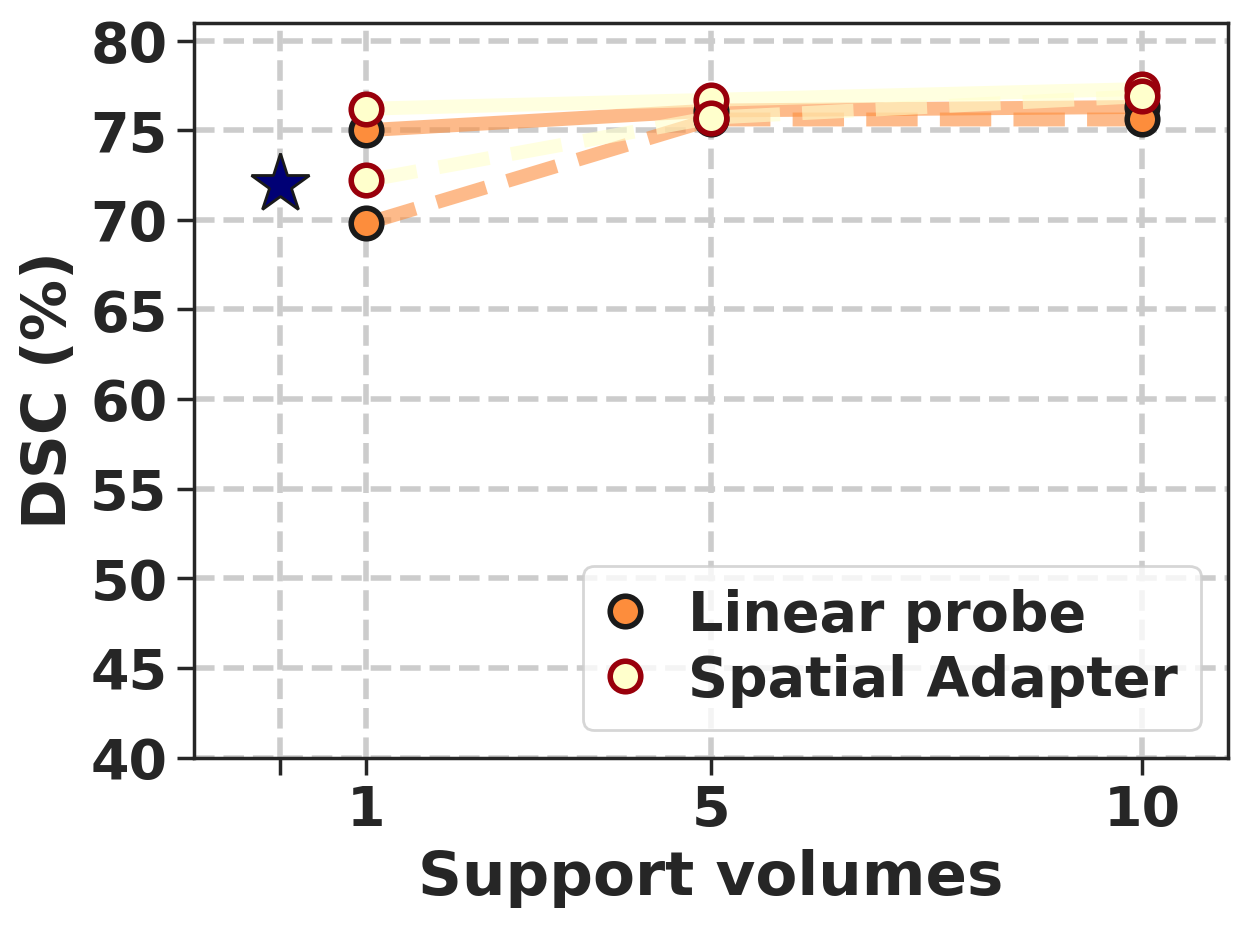}  \\
            \small \textbf{(a) PEFT} & \small \textbf{(b) Black-box}
         \end{tabular}

        \caption{\textbf{Head initialization.} Impact of random head initialization (\textit{dashed lines}) compared to using the pre-trained zero-shot prototypes for each task (\textit{solid lines}). Results averaged across nine base organs on TotalSegmentator.}
        \label{fig:ablation_head}
\end{figure}

\textit{\textbf{The role of the head initialization during adaptation}}. As stated in the implementation details of the adaptation stage (\textit{see} Section~\ref{sec:train_initialization}), the classification head is initialized with the weights and bias obtained for the target organ during pre-training. Although common practice involves randomly initializing the linear classifier weights, recent literature stresses the importance of using a good initialization of this layer for fine-tuning \citep{LPFT} and black-box Adapters \citep{CLAP}. The effect of this methodological choice is studied in Figure~\ref{fig:ablation_head} for PEFT and back-box Adapters. Results highlight the importance of proper head initialization. For instance, LoRA largely degrades its performance if the classification head is randomly initialized, whose impact is consistent regardless of the number of labeled samples.

\begin{figure}[t!]
    \centering    \includegraphics[width=.80\linewidth]{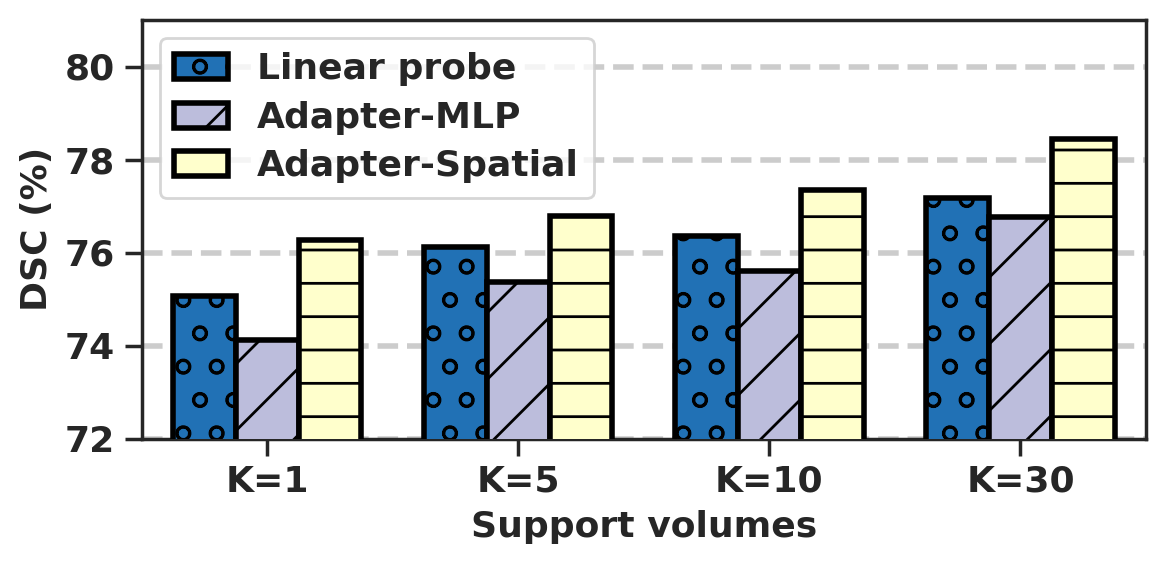}
        \caption{\textbf{Black-box Adapters design}. Effect of including volumetric convolutions on residual Adapters, compared to linear probing or MLP-Adapter \citep{Gao2021}. Results averaged across nine base organs on TotalSegmentator.}
        \label{fig:ablation_adapter}
\end{figure}

\textit{\textbf{On the design of black-box Adapters}}. Figure~\ref{fig:ablation_adapter} compares the proposed Spatial Adapter to relevant black-box baselines: a residual MLP head \citep{Gao2021}, and linear probing. Interestingly, using a set of residual, linear transformations over the pre-trained features as in CLIP-Adapter \citep{Gao2021} provides worse results than a properly initialized linear probe, as recently suggested in \cite{CLAP}. By incorporating spatial information, results consistently improve across the different labeled regimes, which points out the utility of the proposed method in the context of black-box adaptation.

\begin{figure}[t!]
    \centering
            \includegraphics[width=.80\linewidth]{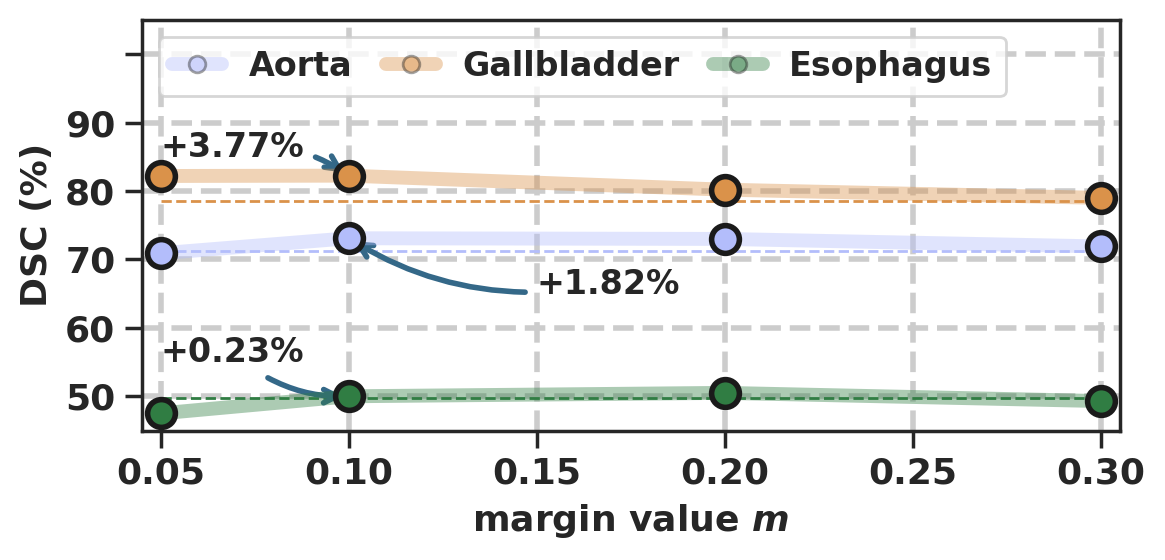}
        \caption{\textbf{Transductive inference configuration}. Study on the hyper-parameter $m$ in Eq. \ref{eq:size_penalty}, which corresponds to the estimated proportion prior margin. Results were obtained using a Spatial Adapter with K=10 support samples for representative structures. Dashed lines indicate the results obtained in the inductive inference, i.e., not incorporating the proportion priors of the query volume.}
        \label{fig:ablation_ti}
\end{figure}

\textit{\textbf{Exploring the transductive inference setting}}. Figure~\ref{fig:ablation_ti} studies the optimum margin for the size regularizer in the transductive setting. Results on three representative structures show that a wide range of $m$ values offer promising results above the baseline. However, a lower margin value ($m=0.05$) might degrade the performance, as the target size is estimated just from a few support samples. Furthermore, an excessively large margin ($m=0.30$) may not affect the query output, as it can result in a significantly relaxed constraint.

%% Conclusions
%-----------
\section{Discussion}
\label{sec:conclusions}

Current trends in computer vision have shown the emergent potential of the so-called foundation models. Pre-trained on large heterogeneous data sources and tasks, these networks can be efficiently transferred to downstream applications. In medical volumetric organ segmentation, the pre-train-and-adapt paradigm is highly desirable due to the burden of annotating scans at the voxel level, as well as the limited data and computational resources of particular clinical institutions. In this context, we have introduced FSEFT. This novel and realistic learning scenario accommodates practical clinical settings, i.e., adapting a large pre-trained model efficiently to a new task/domain with a limited number of labeled samples. To assess this setup, a foundation model for volumetric organ segmentation has been introduced, pre-trained on an assembly of nine open-access datasets, which gathers 29 different anatomical structures and 2,042 partially labeled CT volumes. Next, an efficient transfer learning setting has been described. First, we have assumed a few-shot setting, in which adaptation should be performed using only a few annotated volumes. Second, we argue that the transfer should be performed computationally efficiently, i.e., with minimal computational resources. Two solutions have been explored to address this challenge: PEFT, a set of techniques recently popularized for natural language processing, and black-box adaptation, a privacy-preserving scenario that works directly over feature representations.

We have conducted comprehensive transfer learning experiments with the released foundation model under the FSEFT setting. Such studies include two external datasets, 25 tasks, and several low-shot regimes. These experiments have underscored the limitations of full fine-tuning strategies, which are the leading choice for adapting volumetric medical foundation models \citep{UniversalModel, multitalent,suprem}. Concretely, results demonstrate that popular fine-tuning approaches might distort the rich pre-trained knowledge, as well as being computationally expensive to adapt largely parameterized state-of-the-art segmentation backbones. In contrast, the proposed PEFT methods and black-box Adapters have shown remarkable properties in the FSEFT scenario for both the data and computationally efficient adaptation. We have also provided methodological guidelines for its proper application. These hint at the future significance of supervised pre-training compared to self-supervision, the importance of deploying well-initialized adaptation strategies, the robustness of the proposed black-box Spatial Adapters, and the benefit of prior-aware transductive inference mechanisms. Remarkably, these methods have been successfully applied to foundation models recently published by other authors \citep{UniversalModel,suprem}, manifesting the robustness of the proposed solutions.

Our setting also has potential limitations. First, the transfer of supervised pre-trained foundation models to segment new organs has shown moderate parameter-efficient performance, requiring decoder fine-tuning. Thus, the desired computationally efficient or black-box adaptation is compromised in this scenario. On a positive note, combining PEFT with decoder tuning still provides data-efficient transferability. Unarguably, PEFT methods are more flexible than black-box strategies. The latter purely depends on the output feature representations, whose transfer capabilities will depend on the frequency of the target concept during pre-training, as observed in other foundation models \citep{udandarao2024zeroshot}. This phenomenon may be exacerbated in segmentation tasks due to the decoder specialization. Nevertheless, we would want to note that the merits of one do not detract from those of the other. Instead, they offer orthogonal benefits and adaptation scenarios, i.e., privacy-preserving black-box domain adaptation for base categories, and flexible PEFT when transferring a foundation model to novel tasks. In addition, it is worth noting that the base annotated organs are expected to grow continually since the number of target organs in the human body is naturally finite. Second, this study is limited to CT scans due to the relatively large open-access volumetric labeled data available. Third, we found that PEFT of purely convolutional U-Nets presents limited performance compared to its Transformer-based counterpart. This may be because most recent PEFT methods have been introduced by the natural language community in Transformer blocks, and thus, accommodating PEFT on convolutional architectures requires further studies.

The proposed FSEFT scenario might potentially impact a broad span of clinical applications. Nevertheless, its development in medical image segmentation is in its early stages. This setting has been validated in the context of CT volumetric data, but the key ideas are general to any medical data, such as MRI or 2D image segmentation. In this line, future research to develop modality-specific foundation models for MRI data or assessing the potential inter-modality transfer learning would be of special interest. Also, we want to note that a large body of the pre-training and adaptation techniques is directly imported from other fields such as general vision, natural language processing, or their multimodal intersection. Given the fine-grained specifics of the medical domain, such as inter-annotator uncertainties or large domain shifts in acquisition protocols and scans, appealing future research directions involve exploring the best configuration for PEFT in this medical domain, evaluating the adaptation robustness under domain drifts, or exploring additional priors to include during inference.

\vspace{2mm}
\noindent \textit{\textbf{Acknowledgments}}. This work was funded by the \textit{Fonds de recherche du Québec (FRQ)} and the Natural Sciences and Engineering Research Council of Canada (NSERC). 

%% References
%-------------
%\clearpage
\bibliographystyle{model2-names.bst}\biboptions{authoryear}
\bibliography{refs.bib}

%\clearpage
%\newpage

\end{document}